\documentclass[10pt,twocolumn,letterpaper]{article}

\usepackage{cvpr}              % To produce the CAMERA-READY version
\usepackage{makecell}

\usepackage[dvipsnames]{xcolor}

\usepackage{tikz}
\usepackage{wrapfig}

\definecolor{cvprblue}{rgb}{0.21,0.49,0.74}
\usepackage[pagebackref,breaklinks,colorlinks,citecolor=cvprblue]{hyperref}

\newcommand{\name}{VideoCon\xspace}

\newcommand{\msr}{MSR-VTT\xspace}
\newcommand{\vatex}{VaTeX\xspace}
\newcommand{\tempo}{TEMPO\xspace}
\newcommand{\mplugowl}{mPLUG-Owl-Video\xspace}
\newcommand{\preowl}{\emph{Owl-Base}\xspace}
\newcommand{\ftowl}{\emph{Owl-Con}\xspace}
\newcommand{\randowl}{\emph{Owl-Rand}\xspace}
\newcommand{\pali}{End-to-End VNLI\xspace} % changed by Yonatan
\newcommand{\ex}[1]{``\emph{#1}''}

\def\paperID{8971} % *** Enter the Paper ID here
\def\confName{CVPR}
\def\confYear{2024}

\title{\name: Robust Video-Language Alignment via Contrast Captions}

\author{
Hritik Bansal$^{1}$ 
\quad Yonatan Bitton$^{2}$ 
\quad Idan Szpektor$^{2}$\thanks{Equal Advising.}
\quad Kai-Wei Chang$^{1}$\footnotemark[1] 
\quad Aditya Grover$^{1}$\footnotemark[1]\\
$^{1}$UCLA 
\quad $^{2}$Google Research \\
\texttt{\{hbansal, kwchang, adityag\}@cs.ucla.edu}\\
\texttt{\{yonatanbitton,szpektor\}@google.com}}

\begin{document}

\maketitle
\begin{abstract}
Despite being (pre)trained on a massive amount of data, state-of-the-art video-language alignment models are not robust to semantically-plausible contrastive changes in the video captions. Our work addresses this by identifying a broad spectrum of \textit{contrast} misalignments, such as replacing entities, actions, and flipping event order, which alignment models should be robust against. To this end, we introduce the \name, a video-language alignment dataset constructed by a large language model that generates
plausible contrast video captions and explanations for differences between original and contrast video captions. Then, a generative video-language model is finetuned with \name to assess video-language entailment and generate explanations. Our \name-based alignment model significantly outperforms current models. It 
exhibits a $12$-point increase in AUC for the video-language alignment task on human-generated contrast captions. Finally, our model sets new state of the art zero-shot performance in temporally-extensive video-language tasks such as text-to-video retrieval (SSv2-Temporal) and video question answering (ATP-Hard). Moreover, our model shows superior performance on novel videos and human-crafted captions and explanations. Our code and data are available at \url{https://github.com/Hritikbansal/videocon}.
\end{abstract}

\section{Introduction}
\label{sec:introduction}

Semantically aligning data points from diverse modalities is a long-standing goal of AI. We focus on video-language alignment, which is challenging due to the complexities involved in understanding of entities, their relationships, and temporal order of the depicted events \cite{hendricks2018localizing}. Recent models such as VideoCLIP \cite{xu2021videoclip}, ImageBind \cite{girdhar2023imagebind} learn a shared embedding space. Similarly, generative models such as Flamingo \cite{alayrac2022flamingo}, \mplugowl \cite{ye2023mplug} can provide a classification label (e.g., yes/no) when queried about video-language alignment. 
% The ability to semantically align diverse data modalities is a long-standing goal of AI. Specifically, video-language alignment is challenging due to the complexities involved in understanding of entities, their relationships, and temporal order of the depicted events \cite{hendricks2018localizing}. In this regard, recent instantiations such as VideoCLIP \cite{xu2021videoclip}, ImageBind \cite{girdhar2023imagebind} learn a shared embedding space for video-language input by training on web-scale data such as HowTo100M \cite{miech2019howto100m,xue2022hdvila,grauman2022ego4d,bain2021frozen}. Similarly, there are generative models such as Flamingo \cite{alayrac2022flamingo}, \mplugowl \cite{ye2023mplug} that can generate simple text response (e.g., yes/no) when queried about the alignment between the input data modalities (e.g., does the given video entail `caption'?). Post-training, these models achieve impressive zero-shot performance on downstream tasks such as action recognition \cite{carreira2019short}, retrieval \cite{xu2016msr} and video question answering \cite{xiao2021next}.  

\begin{figure*}[h]
    \centering
    \includegraphics[scale=0.35]{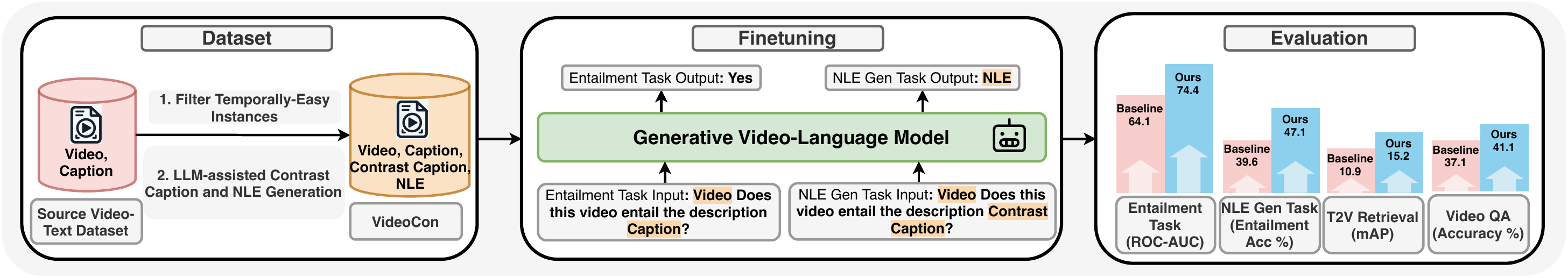}
    \caption{\small{\textbf{
Overview of our \name approach. 
%    Overview of our pipeline which highlights the benefit of finetuning models on the \name data.
    } 
First, aligned video-language pairs are filtered to retain temporally-challenging instances. Then contrast captions and natural language explanations (NLE) are generated by an LLM to create the \name dataset. Second, a video-language alignment model is finetuned with \name on the alignment and NLE  tasks. Finally, the finetuned model is evaluated against the baseline model. Our results show that it  outperforms the baseline, achieving SOTA results on downstream tasks.
%    First, We start with the source video-language datasets which are filtered to retain temporally-rich instances in the dataset. Subsequently, we create the \name dataset by generating contrast captions and natural language explanations (NLE) from a large language model (LLM). Second, we leverage the \name data to finetune a generative video-language model on the video-language entailment task and NLE generation task. Finally, we show that our finetuned model outperforms the pretrained model (baseline) on unseen instances of entailment and NLE generation task. As a result of improved model robustness, we achieve state-of-the-art results on zero-shot downstream tasks such as Text-to-Video (T2V) retrieval and Video QA datasets.
    }}
    \label{fig:pipeline}
\end{figure*}

Despite large-scale pretraining, prior work \cite{park2022exposing,bagad2023test,momeni2023verbs,wang2023paxion} highlights that video-language alignment models are not robust to semantically plausible manipulations to an original aligned caption in the form of contrast captions, such as from `dog runs away \textit{before} it eats food' to `dog runs away \textit{after} it eats food'. Such pitfalls in robustness questions the trustworthiness of alignment models for large-scale deployment. To mitigate these shortcomings, one possible solution is to scale video-language pairs more for increased diversity during pretraining. However, this is challenging due to the difficulties in sourcing new, high-quality and permissible content, as well as the requirement for substantial storage capacity. Several works \cite{gunasekar2023textbooks,gadre2023datacomp,fang2023data} have shown that naively training models on web-scale data has diminishing returns on downstream tasks, and emphasize the importance of data quality. Furthermore, the recent studies \cite{yuksekgonul2022and,li2023desco} demonstrate that applying a contrastive objective to the pretraining datasets does not encourage the model to grasp the fine-grained details within image/region-caption data.

% Despite the large-scale pretraining, prior work \cite{park2022exposing,bagad2023test,momeni2023verbs,wang2023paxion} highlights that these models are not robust to semantically plausible manipulations to the original video captions, in the form of contrast captions. For instance, they cannot distinguish the alignment between a video with the matched caption (e.g.,`two people are \textit{wrestling}') with its contrast version (e.g., `two people are \textit{dancing}'). Such pitfalls in the model's robustness questions their trustworthiness for large-scale deployment in-the-wild. To mitigate such shortcomings, one possible solution is to scale the data more (say $1$B video-language pairs) for increased diversity during pretraining. However, this is challenging due to the difficulties in sourcing new and permissible content from platforms beyond YouTube \cite{zellers2021merlot,abu2016youtube}, as well as the requirement for substantial storage capacity. Several works \cite{gunasekar2023textbooks,gadre2023datacomp,maini2023t,fang2023data} have shown that training models on the passively increasing web-scale data has diminishing returns on downstream tasks, and emphasize instead on the critical importance of data quality. Thus, we adopt a scalable, active strategy to gather high-quality data that is deliberately enriched with the attributes that we want to instill in the models. In this work, we take the later approach, and show that it is effective in training robust video-language alignment models.

\begin{figure*}[h]
    \centering
    \includegraphics[scale=0.3]{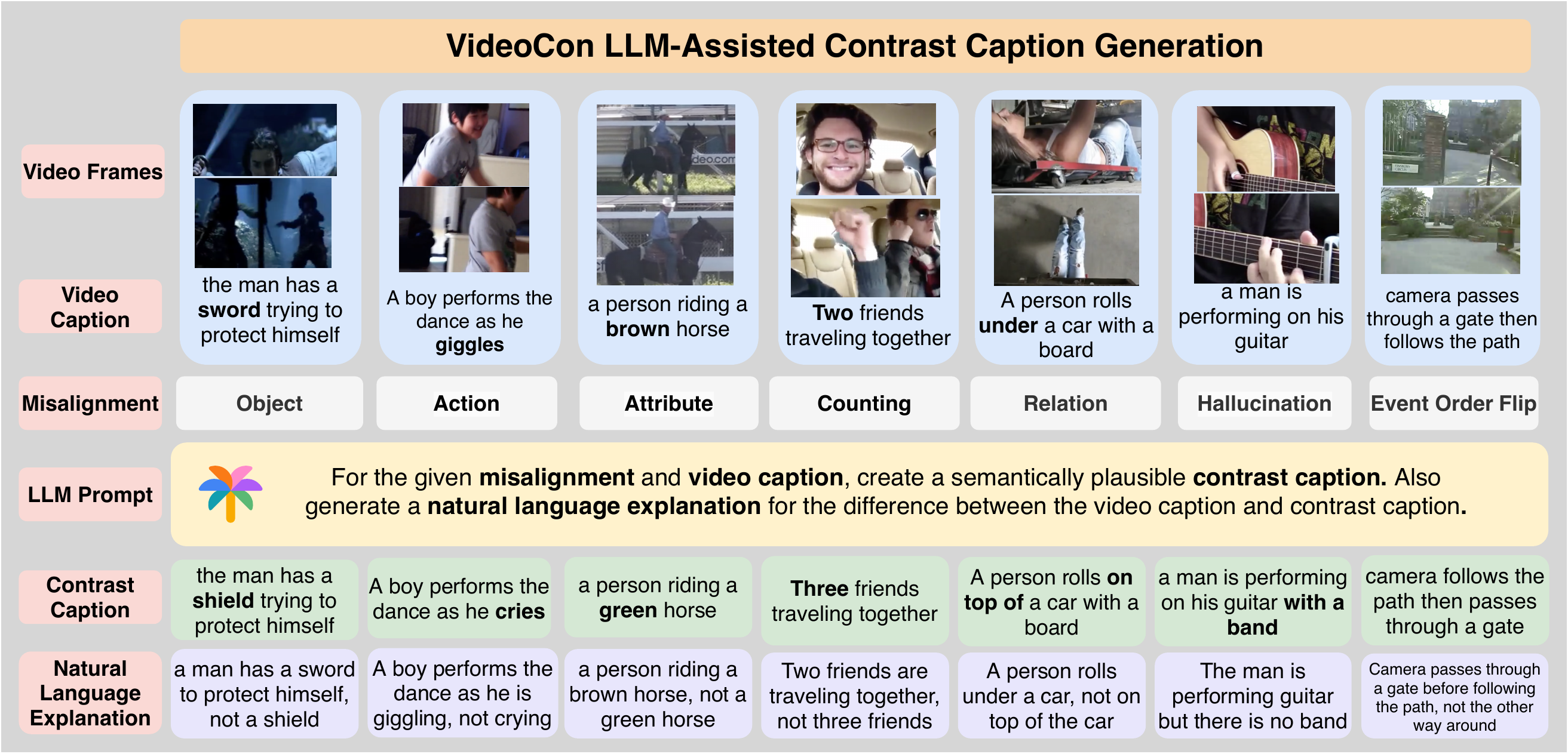}
    \caption{\textbf{Overview of the \name data generation process from top to bottom.} Specifically, we prompt a large language model (PaLM-2) with the original caption that is grounded in the video, and the intended type of misalignment within the contrast caption. We consider \textit{seven} kinds of misalignments including object, action, attribute, counting, spatial relation, hallucination, and event order flip. We provide a generated contrast caption and the corresponding natural language explanation for each misalignment type.}
    \label{fig:data_generation}
\end{figure*}

To this end, we take a scalable, active strategy to gather high-quality data that is deliberately enriched with the attributes that we want to instill in alignment models. We create a novel dataset,
\textbf{\name}, to improve the robustness of models. Specifically, the dataset consists of a variety of semantically plausible
video-language misalignments in contrast captions. These misalignments include altering \textit{objects (entities)}, \textit{actions}, \textit{attributes}, \textit{relations}, \textit{counts}, \textit{event orders}, and introducing \textit{hallucinations} (Figure \ref{fig:data_generation}). To construct \name, a large language model (PaLM-2 API) takes video-caption pairs as input and generates high-quality contrast captions for a given misalignment type. 
To make our dataset temporally-challenging, we skipped ``easy’’ video-caption pairs whose alignment could be inferred based on a single frame (image) understanding \cite{buch2022revisiting,lei2022revealing} (\S \ref{sec:temp_challenging_vl_data}). 
In addition, the LLM generates natural language explanations (NLEs) \cite{sammani2022nlx} to the differences between original and altered captions, which are used for further robust training.
We performed human verification on a sample of \name and found that it is of high-quality. Finally, to evaluate the model's generalization capabilities, we collect human-generated contrast captions and NLEs for the videos sourced from external datasets that did not contribute to \name's development.

We finetuned a generative video-language model  (\mplugowl) on the \name dataset. The trained model surpasses existing video-language alignment models by a large margin on the LLM-generated test set for both video-language alignment and NLE generation tasks. Interestingly, we observed that our finetuned model generalizes to unseen videos and human-generated contrast captions and NLEs, and outperforms the baseline models. For instance, our model's ROC-AUC exceeds the baseline model by $12$ points on the human-generated contrast captions. This indicates that our model has developed a better understanding of the entities, their interactions, action understanding, as well as the temporal order of the events for robust video-language alignment.
%Post-data construction, we finetune a generative video-language model \mplugowl on the original captions, LLM-generated contrast captions and NLEs from the \name dataset. We find that this simple approach allows our model to surpass the existing video-language models by a large margin on the LLM-generated test set for both the video-language entailment and NLE generation tasks. Interestingly, we observe that our finetuned model generalizes to out-of-distribution videos and human-generated contrast captions and NLEs, and outperforms the baseline models (\S \ref{sec: experiments}). For instance, we observe that our model's ROC-AUC exceeds the baseline video-language model by $27$ points on the video-language entailment task on the LLM-generated test set. This indicates that our model has developed a better understanding of the entities, their interactions, action understanding, as well as the temporal order of the events for robust video-language alignment. 

We further assessed the effectiveness of robust training via contrast captions on zero-shot downstream video-language tasks such text-to-video retrieval and video question answering on the temporally-challenging and action-intensive SSv2-Temporal \cite{sevilla2021only} and SSv2-Events \cite{bagad2023test}. Our model achieves state-of-the-art (SOTA) performance, improving on SSv2-Temporal by $4.3$ mAP, SSv2-Events by $3.6$ mAP points. In addition, our model also achieves SOTA on temporal and causal video question answering in the ATP-Hard dataset, increasing $4\%$ accuracy. This suggests that equipping a model with the knowledge of contrast captions is highly data-efficient and effective in improving its robustness in comparison to scaling the pretraining data. The complete pipeline is illustrated in Figure \ref{fig:pipeline}. The dataset and the model
will be released upon acceptance.

% \ag{we should mention if we are going to make our dataset public (if not, then skip)}
%We further assess the effectiveness of robust training via contrast captions on video-language downstream tasks such text-to-video and video question answering. We find that our model achieves state-of-the-art performance on the zero-shot downstream tasks including text-to-video retrieval on temporally-challenging and action-intensive SSv2-Temporal \cite{sevilla2021only} and SSv2-Events \cite{bagad2023test}, and video question answering on temporal-challenging ATP-Hard \cite{buch2022revisiting}. Specifically, our model sets a new SOTA by outperforming the the existing models on SSv2-Temporal by $4.3$ mAP, SSv2-Events by $3.6$ mAP points. In addition, our model also achieves SOTA on temporal and causal video question answering ATP-Hard dataset by outperforming an existing model by $4\%$ accuracy. This suggests that equipping a model with the knowledge of contrast captions is highly data-efficient and effective in improving its robustness in comparison to scaling the pretraining data. The complete pipeline of our work is illustrated in Figure \ref{fig:pipeline}.

% We summarize our contributions as:

% \yb{consider 1. data generation method, including the misalignments + filtering techniques; 2. the published dataset, including train/test, auto-generated + human-collected test, high quality via human verification.
% 3. model}\hb{will add later}

\section{Video Language Alignment}
\label{sec:task}

We are interested in assessing the semantic alignment between the video\footnote{Like prior works \cite{xu2021videoclip,luo2022clip4clip}, we use  only the video frames (the visual channel) without the soundtrack (the audio channel).} and text data since it powers many practical applications such as video-text retrieval \cite{xu2016msr}, video generation \cite{blattmann2023align,wang2023modelscope} and video captioning \cite{yang2023vid2seq}.
%We are interested in assessing the semantic alignment between the video \footnote{Like prior works \cite{xu2021videoclip,luo2022clip4clip}, we use the video frames (the visual channel) and not the soundtrack (the audio channel).} and text data since it powers many practical applications such as video-text retrieval \cite{xu2016msr}, video generation \cite{blattmann2023align,wang2023modelscope} and video captioning \cite{yang2023vid2seq}. 
To this end, \cite{xu2021videoclip,girdhar2023imagebind,wang2022internvideo,radford2021learning} designed (image)video-text alignment models that are utilized for evaluating the semantic similarity between the two modalities. However, previous works \cite{park2022exposing,momeni2023verbs,bagad2023test,wang2023paxion} have questioned their robustness to semantically plausible changes to the video descriptions, termed here \emph{contrast captions}. Our aim is to improve the robustness of video-text alignment models by training on contrast captions with a wide range of misalignments.
%To this end, \cite{xu2021videoclip,girdhar2023imagebind,wang2022internvideo,radford2021learning} designed (image)video-text alignment models that are currently utilized for evaluating the semantic closeness between the two modalities. However, \cite{park2022exposing,momeni2023verbs,bagad2023test,wang2023paxion} have questioned their robustness to semantically plausible changes to the video descriptions (contrast captions). Hence, our aim is to build a robust video-text alignment model by training on a wide range plausible misalignments in the contrast captions.

Consider a dataset $\mathcal{D}=\{(V_i, T_i, C_i, E_i)\}$ where $V_i$ is a video, $T_i$ is an aligned caption, $C_i$ is a contrast caption which is
a perturbation of $T_i$ but misaligns with $V_i$, and $E_i$ is a natural language explanation for the misalignment between $V_i$ and $C_i$. We consider two video-language alignment tasks: (a) video-language entailment, (b) natural language explanation. 

\paragraph{Video-Language Entailment (VLE)} casts video-text alignment as a Visual Entailment (VE) task. VE was originally defined for images as premises and texts as hypothesis  \cite{xie2018visual,xie2019visual}. We extend VE definition also for videos as premises, under  which a classification model $A_{vle}(V, T)$ predicts whether a video $V$ entails a text $T$. %\paragraph{Video-Language Entailment (VLE)} involves a model, $A_{en}$, scoring the alignment between video and text from 0 to 1, aiming for perfect alignment ($A_{en}(V, T) = 1$) with video-text pairs and misalignment ($A_{en}(V, C) = 0$) with video-contrast pairs. Leveraging video-language representation learning, the model learns to distinguish between matched $(V, T)$ and unmatched $(V, C)$ pairs, potentially using a multimodal generative model to affirm or deny entailment with simple responses to queries about the video-text relationship.

\paragraph{Natural Language Explanation (NLE)} requires a model, $A_{nle}(V, C)$, to generate an open-ended explanation for the discrepancy between a video $V$ and a non-entailing caption $C$.
%\paragraph{Natural Language Explanation (NLE) Generation} requires a model, $A_{exp}$, to articulate the discrepancy between a video (V) and an inaccurate caption (C). Once the VLE model identifies a general misalignment, $A_{exp}$ is used to elucidate the specifics, generating an explanation ($\hat{E}$) closely mirroring the original NLE ($E$). 

In this paper, we address both VLE and NLE tasks under a multitask setting in which a single video-language generative model generates the binary label for entailment and the open-ended explanation.
%Here, a unified video-language generative model adept at both close-ended (binary e.g., `Yes' or `No') and open-ended (descriptive) queries is employed to perform both VLE and NLE tasks effectively.

% \yb{---TO HERE}

% In our work, we employ a unified video-language generative model to realize the role of $A_en$ and $A_exp$ since is adept at handling both close-ended queries (e.g., expected response is binary `Yes' or `No') and open-ended queries (e.g., describing the actions in a video).
\section{\name: Contrast Captions Generation for Robust Video-Language Alignment}
\label{sec:dataset_construction}

Our research goal is to measure the impact of a comprehensive dataset on increasing the robustness of video-text alignment models. To this end, we first
collect video-caption pairs where the caption cannot be derived from a single frame of video. We then categorize a wide range of semantically plausible manipulations of video captions. Using an LLM for large-scale computation, contrast captions and related explanations are generated for the defined categories, constructing the \name dataset. Finally, we extend \name to include human-created contrast captions as held-out evaluation on unseen videos. We detail the dataset construction steps below.
%Here, our goal is to construct a comprehensive dataset for training robust video-language alignment models. To this end, we categorize a wide range of plausible semantic manipulations for the video descriptions. Firstly, we perform dataset filtering using \pali (visual entailment) model \cite{yarom2023you} to retain temporally-challenging instances from the video-text datasets. Subsequently, we utilize LLM for large-scale and cost-effective generation of the contrast captions and the natural language explanations for a matched video-text pair from existing datasets. Finally, we extend \name to perform generalization evaluation of the trained models on unseen videos and \textit{human}-generated contrast captions and natural language explanations. We describe the dataset construction pipeline here.

\subsection{Temporally-Challenging Instance Selection}
\label{sec:temp_challenging_vl_data}

% \yb{Consider making this subsection shorter. The summary/TLDR is around `We refined the \msr, \vatex, and \tempo datasets to enhance temporal complexity, using the PaLI-17B model to filter out 'temporally-easy' video-text pairs with high image-text alignment scores. Post-filtering, the share of 'temporally-challenging' instances increases to 81.5\% in \msr and 71\% in \vatex, better serving video-language model training.`}

To construct \name, we start with existing datasets that include natural (real) videos and associated high-quality human-written captions: \msr \cite{xu2016msr}, \vatex \cite{Wang_2019_ICCV}, and \tempo \cite{hendricks2018localizing}. \msr and \vatex consist of $20$ captions and $10$ captions per video, respectively, while \tempo consists of a single caption per video. 
More dataset details are in Appendix \S \ref{sec:details_vld}. 
%We annotate the existing datasets comprising of natural (real) videos and associated high-quality human-written captions. Specifically, we include \msr \cite{xu2016msr}, \vatex \cite{Wang_2019_ICCV}, and \tempo \cite{hendricks2018localizing} video-language datasets. \msr and \vatex consists $20$ captions and $10$ captions per video, respectively. \tempo consists of single caption grounded in the video. It is specifically designed to create temporally-challenging instances. We add more details about the datasets in Appendix \S \ref{sec:details_vld}. 

\tempo is designed to create temporally-challenging instances, while \msr and \vatex contain more general video-caption pairs.
For \msr and \vatex, we filter out instances, where the caption is highly associated with a single frame in the video based on an image-text alignment model. In such cases, a video-text alignment can leverage shortcuts and align the video to its caption without understanding the temporal or causal relations depicted in the video. We want to filter such instances.

To this end, we employ the \pali model \cite{yarom2023you} to calculate an alignment score $A_{vle}(V, T)$ between a video $V = \{I_1, I_2, \ldots, I_N\}$ and a text $T$ where $I_i$ is a frame from the video sampled at a rate of 1 frame per second. Formally, 
\begin{equation}\label{eq:vnli}
    A_{vle}(V, T) = \text{max}_{i}(VNLI(I_i, T))
\end{equation}
where $VNLI(I_i, T)$ is the image/text entailment score. 
% Each $\{V, T\}$ pair is  out as ``easy’' if $A_{vle}(V, T) > 0.5$. 
There are 20 and 10 captions per video in the \msr and \vatex datasets, respectively. We retain 5 captions per video from these datasets with the lowest $A_{vle}(V, T)$, and the remaining captions are filtered out. Post-filtering, the percentage of temporally-challenging instances increased from $36.5\%$ to $81.5\%$ in \msr, and from $42.6\%$ to $71\%$ in \vatex.

\subsection{Categories of Contrast Captions}
\label{sec:misalignment_types}

We aim for \name to include a wide range of misalignments in its contrast captions. Overall, \name covers \textit{seven} misalignment types, exemplified in Figure \ref{fig:data_generation}. We include replacement of \textit{objects} (entities) and \textit{actions} following the analysis in \cite{park2022exposing,momeni2023verbs}, and replacement of  \textit{attributes}, \textit{counts}, \textit{relations}, as well as adding unrelated but plausible information to captions as \textit{hallucinations} following \cite{li2023evaluating,ma2023crepe,lu2023evaluation}’s study of image/text alignment model brittleness. Since most video-text models rely on pretrained image backbones, they are likely to suffer from similar problems. Finally, following \cite{bagad2023test}’s analysis that video-text models do not understand temporal order of the events, we include \textit{event order flipping} as misalignment type.

\subsection{Data Generation using an LLM}
\label{sec:llm_generated}

To generate contrast captions and corresponding NLE we first assign one of the seven misalignment types (\S \ref{sec:misalignment_types}) to each caption in the input video-text datasets (\S \ref{sec:temp_challenging_vl_data}) (details in Appendix \S \ref{sec:misalignment_assignment}). Then, given a video $V$ and a misalignment type $m$, we prompt PaLM-2 API\footnote{\url{https://developers.generativeai.google/products/palm}} \cite{anil2023palm} to generate a contrast caption and accompanied explanation (our type-specific prompts are detailed in Appendix \S \ref{sec:llm_prompt}).
%Here, we aim to generate the contrast captions and corresponding natural language explanation (NLE) for their difference with the expected caption using an LLM $L$. Specifically, we assign one of the seven misalignment types to each caption from the video-text datasets (Appendix \S \ref{sec:misalignment_assignment}). Following \cite{yarom2023you,momeni2023verbs}, we use PaLM-2 API \cite{anil2023palm} for data generation such that $C, E \sim L(T, m)$ where $C$ and $E$ are the generated contrast caption and natural language explanation $E$ for the video caption $T$ and assigned misalignment $m \in $ \{object, action, attribute, count, relation, hallucination, and flipping event order\}. We provide the LLM prompts in App. \S \ref{sec:llm_prompt}. 

% \begin{wrapfigure}{r}{0.6\linewidth}
%     \includegraphics[width=\linewidth]{images/data_distribution.pdf}
%     \caption{\textbf{Distribution of the types of misalignments within the contrast captions of the \name dataset.} We observe that the dataset has good representation for all the kinds of misalignments from $8.8\%$ to $24.2\%$.}
%     \label{fig:misalignment_distribution}
% \end{wrapfigure}

\begin{figure}
    \centering
    \includegraphics[scale=0.55]{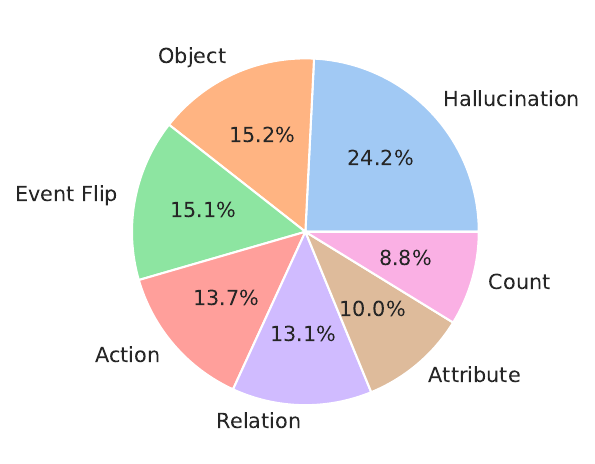}
    \caption{\textbf{Distribution of the types of misalignments within the contrast captions of the \name dataset.} We observe that the dataset has good representation for all the kinds of misalignments ranging from $8.8\%$ to $24.2\%$.}
    \label{fig:misalignment_distribution}
\end{figure}

% An example for the generated data is:

% \noindent \textcolor{blue}{\textbf{Input to LLM}}:\\
% \indent \textcolor{blue}{Misalignment}: \textit{Event Order Flipping}\\  
% \indent \textcolor{blue}{Original Video Caption}: \textit{A boy starts a countdown and then starts eating chips.}\\  
% \textcolor{violet}{\textbf{LLM output}}:\\
% \indent \textcolor{violet}{Contrast Caption}:  \textit{A boy starts eating chips and then starts a countdown.}\\    
% \indent \textcolor{violet}{Natural Language Explanation}: \textit{A boy starts a countdown before eating chips, not after eating chips.}\\

% For instance, bad contrast captions will still entail the video, and some of the NLE provide insufficient description about the discrepancy in the video caption and contrast caption. 

Analyzing the LLM generations, we found that sometimes the output caption $C$ do not contradict the original caption $T$. For example, a generated contrast caption \ex{a person riding a car} does not contradict the original caption \ex{a person riding a mustang}.
To filter such cases, we employ a Natural Language Inference (NLI) model \cite{honovich2021q} and remove cases in which the contrast caption is assessed as entailed by the original caption $NLI(T, C) > 0.5$. 
Post-filtering, each tuple $(V, T, C, m)$ is converted to the two instances of video/language entailment task: $A_{en}(V, T) = 1$ and $A_{en}(V, C) = 0$. We present the dataset statistics for the entailment task in Table \ref{tab:data_stats}, including train/eval/test splits. In addition, Fig. \ref{fig:misalignment_distribution} shows the distribution of misalignment types in the dataset. We observe that \name maintains a high density across the $7$ misalignments ranging from $8.8\%$ to $24.2\%$.
%In some cases, we find that the LLM-generated contrast caption or NLE are not accurate. To filter such instances, we use an English natural language inference model \cite{honovich2021q} $NLI({s}_{1}, {s}_{2})$ to provide the entailment score $[0-1]$ between the premise ${s}_{1}$ and hypothesis ${s}_{2}$. Specifically, we filter the generations where $NLI(T, C) > 0.5$ which removes the cases where the contrast caption entails the original caption. For example, a generated contrast caption `a person riding a car' does not contradict the original video caption `a person riding a mustang'. Post-filtering, each tuple $(V, T, C, m)$ can be converted to the two instances of entailment task i.e., $A_{vle}(V, T) = 1$ and $A_{vle}(V, C) = 0$. We present the dataset statistics for the video-language entailment task in Table \ref{tab:data_stats}. In addition, we show the distribution of various misalignments $m$ in the entire entailment dataset in Figure \ref{fig:misalignment_distribution}. We observe that the \name dataset maintains a high density across the various plausible misalignments in the contrast captions ranging from $8.8\% - 24.2\%$.

We also found that some generated explanations do not describe the differences between $T$ and $C$ well. For example, the explanation \ex{two friends are not traveling together} does not fully describe the discrepancy between \ex{three friends traveling together} and \ex{two friends are traveling together}.
To filter these out, generated examples are removed if $NLI(F(T, C), E) < 0.6$ where $F(T, C)$ is the premise comprising the original and contrast captions. Specifically, premise will be `Expected Caption: $T$ Actual Caption: $E$' and hypothesis will be `Difference between Expected and Actual Caption: $E$'. This filter indicates that the information in the explanation is not entailed by the difference between the two captions. The dataset statistics for the NLE task is presented in Table \ref{tab:data_stats}. We refer to the final LLM-generated dataset as \name (\textbf{LLM}).
%To filter for the NLE task, we start with the filtered entailment data and remove the generated instances $NLI (F(T, C), E) < 0.6$ where $F(T, C)$ is the premise comprising the expected and actual video caption. This filter indicates that the generated NLE does not entail the difference between the original and contrast caption. For example, a NLE `two friends are not travelling together' does not fully describe the discrepancy between the expected text grounded in the video `three friends travelling together' for the actual text `two friends are travelling together'. We present the dataset statistics for the video-language NLE generation task in Table \ref{tab:data_stats}. We refer the LLM-assisted generated data as \name (\textbf{LLM}).

\begin{table}[t]
\resizebox{\linewidth}{!}{%
\begin{tabular}{l|rrr|rrr}
\hline
        & \multicolumn{3}{c}{\textbf{Video-Language Entailment}} & \multicolumn{3}{|c}{\textbf{Natural Language Explanation}} \\\hline
\multicolumn{1}{c|}{\textbf{Source}} & \multicolumn{1}{c}{\textbf{Train}}       & \multicolumn{1}{c}{\textbf{Val}}       & \multicolumn{1}{c|}{\textbf{Test}}      & \multicolumn{1}{c}{\textbf{Train}}   & \multicolumn{1}{c}{\textbf{Val}}     & \multicolumn{1}{c}{\textbf{Test}}    \\\hline
MSR-VTT & 38366       & 478       & 16538     & 15888    & 206     & 6788    \\
VaTeX   & 66480       & 736       & 8110      & 30180    & 345     & 3636    \\
TEMPO   & 10712       & 7098      & 2708      & 4165     & 2739    & 1073    \\\hline
Total   & 115558      & 8312      & 27356     & 50233    & 3290    & 11497  \\\hline
\end{tabular}%
}
\caption{Statistics for the VLE and NLE tasks in \name.}
\label{tab:data_stats}
\end{table}

To assess the quality of \name (\textbf{LLM}), we perform human evaluation on $500$ contrast captions and NLEs (details in Appendix \ref{sec:human_annotation_data_quality}). The human evaluator found $91\%$ of the contrast captions and $89\%$ of the NLEs to be valid, indicating the high-quality of \name (\textbf{LLM}).
%To assess the quality of our generated dataset, we perform human evaluation on the validity of $500$ contrast captions and NLEs. We provide more details about the human annotation process in . We find that $91\%$ of the contrastive captions and $89\%$ of the NLEs are deemed valid by the human evaluators. It indicates the high-quality of the dataset generated by the LLM.

\subsection{Data Generation using Humans}
\label{sec:data_gen_human}

To study whether a model trained on \name (\textbf{LLM}) generalizes to out-of-distribution videos and its performance on human-generated contrast captions, we randomly selected a set of videos from the validation set of ActivityNet \cite{caba2015activitynet}. This dataset consists of captions matched with segments in the video, e.g., \ex{a little boy is climbing on an outside gym} matched to the first 10 seconds of its related video. We extracted video segments with an associated caption. Human workers\footnote{A shortlist that passed our qualification test.} on Amazon MTurk were then shown the video segments and associated captions and were asked to create a semantically plausible contrast caption and a corresponding NLE (more details in Appendix \S \ref{sec:human_data_collection}). \textit{We did not communicate any type of target misalignments to encourage natural diversity of human created contrast captions}. 
%Here, we seek to study whether the model training on LLM-generated contrast captions generalizes to out-of-distribution videos and the human-generated contrast captions. To this end, we randomly select a set of videos from the validation set of ActivityNet \cite{caba2015activitynet} video-language dataset. The dataset consists of the captions for the segments in the video (e.g., `A little boy is climbing on a outside gym' is matched to first 10 seconds of the video). We first preprocess the data to extract individual video segments with an associated caption. We shortlist human workers that passed our qualification test on the Amazon MTurk platform. Finally, the human workers were shown the video segment and its associated caption, and asked to create a semantic plausible contrast caption along with a natural language explanation for the difference between the video caption and their written contrast caption. We \textbf{do not} communicate the type of misalignments in our LLM-generated contrast captions to encourage diversity in the human created contrast captions (Appendix \S \ref{sec:human_data_collection}). 

Overall, we collected $570$ 
tuples $(V, T, C_{human}, E_{human})$ where $V$ is the video, $T$ is the original caption, $C_{human}$ is the human-written contrast caption, and $E_{human}$ is the human-written explanations. We denote this dataset by \name (\textbf{Human}). We sample $100$ instances from this dataset, and found $93\%$ to be clean. In addition, we observe that many of the human-generated contrast captions perturbing one or more objects ($35\%$) and actions ($35\%$) depicted in the caption. While $8\%-10\%$ of the contrast captions flip the order of the events and attribute of the objects. As this dataset is largely unfiltered, it contains a mix of temporally-easy and challenging instances. We also constructed a more temporally-challenging subset of $290$ instances, denoted \name (\textbf{Human-Hard}), by filtering out tuples in which $A_{vle}(V,T) < 0.5$ (Eq.~\eqref{eq:vnli}), as in \S \ref{sec:temp_challenging_vl_data}. 

%Overall, we collected $570$ instances of a tuple $(V, T, C_{human}, E_{human})$ where $V$ is the video, $T$ is the associated caption, $C_{human}$ is the human-written negative caption, and $E_{human}$ is the human-written explanations. We label this data as \name (\textbf{Human}). As this dataset is largely uncurated, it contains a mix of temporally-easy and challenging instances. To make the dataset more video-oriented, we select a subset instances (tuples) for the which $A_{vle}(V,T) < 0.5$ using the \pali model as described in Eq.~\eqref{eq:vnli}. As a result, we have a subset of $290$ instances, which we label as \name (\textbf{Human-Hard}). 

\section{Experimental Setup}
\label{sec:setup}

We next describe our evaluation setting for measuring the impact of \name on video-text alignment modeling.
%Here, we utilize the \name (LLM) dataset for robust training. We provide the details about the model, task prompts (\S \ref{sec:model}), and evaluation (\S \ref{sec:evaluation}). We also present the list of baseline models in \S \ref{sec:baselines}. 

\subsection{Finetuning with \name}
\label{sec:model}

Our goal in constructing \name (\textbf{LLM}) is to improve robustness of video-text alignment models by fine-tuning on this dataset. To this end, we start with the \mplugowl model \cite{ye2023mplug}, denoted \preowl. Its building blocks are CLIP \cite{radford2021learning} as visual encoder and LLaMA-7B \cite{touvron2023llama} as text encoder/decoder and it was pretrained on VideoChat \cite{li2023videochat}.
%Post-data creation, we utilize \name for training robust video-language alignment model. Specifically, we finetune a pretrained generative video-language model \mplugowl $p_\theta$ on the dataset \cite{ye2023mplug}. The model is pretrained on the VideoChat data \cite{li2023videochat}, which allows it to understand video-centric instructions (in text) and generate text response. Specifically, the model consists CLIP \cite{radford2021learning} as the visual encoder, and LLaMA-7B \cite{touvron2023llama}. 

\begin{figure}[h]
\centering
\resizebox{!}{0.37\linewidth}{%
\begin{tikzpicture}
    \node[draw, rounded corners, fill=blue!10,align=left] {\underline{\textbf{Entailment Task}}: \\ \\
    \textbf{Given:} 
    \textbf{V} (Video), \textbf{T} (Caption), \textbf{C} (Contrast Caption) \\ \\ 
    \textbf{Instruction (I):}  \textit{[\textbf{V}] Does this video entail the description [\textbf{T}]?} \\
    \textbf{Response (R):} \textbf{\textit{Yes}} \\ \\ 
    \textbf{Instruction (I):} \textit{[\textbf{V}] Does this video entail the description [\textbf{C}]?} \\
    \textbf{Response (R):} \textbf{\textit{No}}
    };
\end{tikzpicture}%
}
\caption{Entailment task prompt for finetuning.}
\label{tikz:1}
\end{figure}

\begin{figure}[h]
\centering
\resizebox{!}{0.32\linewidth}{%
\begin{tikzpicture}
    \node[draw, rounded corners, fill=blue!10,align=left] {\underline{\textbf{Natural Language Explanation Generation Task}}: \\ \\
    \textbf{Given:} 
    \textbf{V} (Video), \textbf{C} (Contrast Caption), \textbf{E} (NLE)   \\ \\ 
    \textbf{Instruction (I):} \textit{[\textbf{V}] What is the misalignment between this} \\\textit{ video and the description [\textbf{C}]?} \\
    \textbf{Response (R):} \textit{[\textbf{E}]}
    };
\end{tikzpicture}%
}
\caption{NLE generation task prompt for finetuning.}
\label{tikz:2}
\end{figure}

\begin{table*}[h]
\centering
\begin{tabular}{lccc}
\hline
\textbf{Models}               & \textbf{\name (LLM) Test} & \textbf{\name (Human)} & \textbf{\name (Human-Hard)} \\
\hline
Random & 50.0 & 50.0 & 50.0 \\
\hline
VideoCLIP \cite{xu2021videoclip} & 53.2 & 47.3 & 47.5\\
ImageBind (Video-Text) \cite{girdhar2023imagebind} & 57.1     & 65.2       & 63.0              \\
\preowl \cite{ye2023mplug}        & 57.2    & 66.8       & 64.1            \\
\randowl     &   59.7 &   68.9     & 65.5   \\
\pali \cite{yarom2023you}        & 67.0       & 72.4       & 65.0              \\
\ftowl (\textbf{Ours}) & \textbf{84.6}    & \textbf{78.3}       & \textbf{74.4}           \\\hline
\end{tabular}
\caption{\small{ROC-AUC scores of the tested models for the entailment task on \name test sets.}}
\label{tab:vc_entailment}
\end{table*}

To fine-tune \preowl on \name (\textbf{LLM}), its $\{V, T, C, E\}$\footnote{V: video, T: original caption, C: contrast caption, E: explanation.} tuples were converted into two types of multimodal instruction-response pairs, one for the VLE task $(I_{vle}, R)$ (Fig. \ref{tikz:1}) and one for the NLE task $(I_{nle}, R)$ (Fig. \ref{tikz:2}).
We then train \preowl on all instruction pairs from both the tasks with maximum likelihood loss, resulting in a single model \ftowl. 

%For finetuning, we convert the instances in the \name dataset into multimodal instruction-response $(I, R)$ pairs to perform both the entailment and NLE generation tasks (\S \ref{sec:task}). We finetune the model to increase the likelihood of the response $R$ for the instruction $I$ i.e., $\theta^{*} = argmax_\theta(p_\theta(R | I))$.

%In Figure \ref{tikz:1}, we show the conversion of an instance $(V, T, C)$ into two instruction-response pairs for the entailment task. Here, the instruction $I = M(V, Y)$ where $M$ is the entailment template and $Y \in \{T, C\}$ is the original or contrast caption. Similarly, we illustrate the mapping of an instance $(V, C, E)$ into a single instruction-response pair for the NLE task in Figure \ref{tikz:2}. Here, the instruction $I = N(V, C)$ where $N$ is the NLE task template. We present the finetuning setup in Appendix \S \ref{sec:finetuning}.

\subsection{\name Evaluation Metrics}
\label{sec:evaluation}

To evaluate the performance of the \ftowl on video-text alignment we generate $\ftowl$ response to prompt $I_{vle}$ for video $V$ and text $Y \in \{T, C\}$. We then calculate the probability of generating responses $s_{y} = $\ftowl$(\text{`Yes'}|I_{vle}(V, Y))$ and $s_{n} = $\ftowl$(\text{`No'}|I_{vle}(V, Y))$, and based on these scores the probability for class ‘Yes’:
 $\label{eq:entailment}
P_{yes}(V, Y) = \frac{s_{y}}{s_{y} + s_{n}}    
$. Finally, we compute the ROC-AUC score for $P_{yes}(V, Y)$ over the \name (\textbf{LLM}) eval set, with $\{V, T\}$ as label $1$ and $\{V, C\}$ as label $0$.
%Here, we evaluate the performance of the \mplugowl finetuned (ft.) with the \name model on the entailment and NLE task. To utilize the generative model $p_{\theta^{*}}$ for the entailment task $A_{en}$, we prompt it with the query instruction $I$ consisting of video $V$ and the text $Y \in \{T, C\}$. We calculate the probability of the response $s_1 = p_{\theta^*}(\text{`Yes'}|M(V, Y))$ and $s_2 = p_{\theta^*}(\text{`No'}|M(V, Y))$. Finally, we calculate the entailment score between the video and the text as $\label{eq:entailment}
%A_{en}(V, Y) = \frac{s_1}{s_1 + s_2}    
%$.
% \begin{equation}\label{eq:entailment}
% A_{en}(V, Y) = \frac{s_1}{s_1 + s_2}    
% \end{equation}
%Finally, we calculate the ROC-AUC score for the entailment scores $A_{en}(V, Y)$ given to the instances from the evaluation set. Specifically, the original video captions are considered as label $1$ and contrast captions are considered as label $0$ while computing the ROC-AUC score.

To evaluate \ftowl on the NLE task, we prompt it with instruction $I_{nle}$ instantiated on $\{V, C\}$ pairs from the \name (\textbf{LLM}) eval set. We compare the generated explanation $\hat{E}$  to the ground truth $E$ by measuring entailment probability $NLI(E, \hat{E})$. In our experiments, we experiment with two $NLI$ automatic metrics: (a) $Q^2$ score \cite{honovich2021q}, and (b) PaLM-2 API. We performed human evaluation to measure the agreement between the automatic metrics and the human-rating. We found that both metrics achieve high agreement with human assessment (Appendix \S \ref{sec:human_agreement_nle}).

%To utilize the generative model for the NLE task $A_{exp}$, we prompt it with the query instruction $I$ consisting of the video $V$ and the contrast caption $C$ from the evaluation set. We generate the NLE following $\hat{E} \sim p_{\theta^{*}}(.|I)$ where $I = N(V, C)$. Finally, we calculate the closeness between the generated NLE $\hat{E}$ and the ground-truth NLE $E$ by posing it as textual entailment task $NLI_{Text}(H, P)$ where the hypothesis $H = (T, C, E)$ is the information about the original video caption, contrast caption, and the ground-truth NLE in the evaluation set, and the premise $P = \hat{E}$. In our experiments, we experiment with two $NLI_{text}$ models: (a) $Q^2$ \cite{honovich2021q} that returns an entailment score between $0-1$, and (b) PaLM-2 API that returns $1$ for entailment and $0$ for contradiction. In our experiments, we also perform human evaluation to measure the agreement between the automatic metrics and the human-rating (\S \ref{sec:exp_nle}). We find that both metrics achieve high agreement with human quality assessment.

\subsection{Video-Text Downstream Tasks}
\label{sec:downstream_setting}

We complement the \name intrinsic evaluation over the testset with an extrinsic evaluation over two temporal and action difficult downstream tasks.

We evaluate alignment model performance for \textit{text2video retrieval} over SSv2-Temporal \cite{sevilla2021only} and SSv2-Events \cite{bagad2023test} datasets. We consider the SSv2-Template captions instead of the label captions since they remove the object-centric bias in model evaluation \cite{lei2022revealing}.
We compute input-text/candidate-video alignment score, rank videos and report \textit{mean Average Precision} (mAP). We evaluate alignment model performance for \textit{video question answering} over the ATP-Hard \cite{buch2022revisiting} dataset. We cast each question/candidate-answer pair as an imperative statement using PaLM-2 API, measure alignment to the input video and report \textit{Accuracy}.
More details on the downstream datasets and the evaluation setup are in Appendix \S \ref{sec:dets_downstream}.

\subsection{Baselines} 
\label{sec:baselines}

For the video-text alignment text, we compare \ftowl with the following baselines: (a) \pali as zero-shot \textit{atemporal} model since it does not have access to the temporal order of the video frames, (b) VideoCLIP \cite{xu2021videoclip}, (c) ImageBind \cite{girdhar2023imagebind}, (d) \preowl, and (e) \randowl: \preowl fine-tuned on \name tuples $\{V, T, \hat{C}, E\}$ where $\hat{C}$ is randomly selected from other captions in the dataset.
\randowl would indicate if there is merit in the contrast, hard-negative captions in \name. We include additional baselines TACT \cite{bagad2023test} and VFC \cite{momeni2023verbs} for evaluating on the downstream tasks (\S \ref{sec:downstream}). 
%We include \pali as zero-shot \textit{atemporal} model baseline for the video-text entailment task (\eqref{eq:vnli}). In addition, we consider VideoCLIP \cite{xu2021videoclip}, ImageBind \cite{girdhar2023imagebind}, and pretrained \mplugowl \cite{ye2023mplug} as the relevant baselines for the entailment task. To understand the benefits of the high-quality \name data, we finetune \mplugowl for the entailment task by setting randomly selected original captions from \name (LLM) as the contrast captions (\mplugowl ft. Random). Finally, we include additional baselines such as TACT \cite{bagad2023test} and VFC \cite{momeni2023verbs} for the downstream tasks (\S \ref{sec:downstream}). 

\section{Experiments}
\label{sec: experiments}

\begin{table*}[h]
\centering
\resizebox{\textwidth}{!}{%
\begin{tabular}{lcccc}
\hline
               & \multicolumn{2}{c}{\textbf{\name (LLM)}}                               & \multicolumn{2}{c}{\textbf{\name (Human)}}                             \\\hline
\textbf{Models}         & \makecell{\textbf{$Q^2$ entailment}} & \makecell{\textbf{PaLM-2 entailment acc.} (\%)} & \makecell{\textbf{$Q^2$ entailment}} & \makecell{\textbf{PaLM-2 entailment acc.}(\%)} \\\hline
\preowl  & 0.19                     & 36.8                         & 0.23                     & 39.6                         \\
\ftowl (\textbf{Ours}) & \textbf{0.50}                     & \textbf{65.4}                        &\textbf{ 0.32}                     & \textbf{47.1}              \\\hline          
\end{tabular}%
}
\caption{\small{Performance of the tested models on the NLE generation task, measured via entailment metrics.}}
\label{tab:vc_nle}
\end{table*}

We present our intrinsic (\name eval set) and extrinsic (downstream  tasks) evaluation results, showing the benefits of \name for robust video-language alignment. 
%We show that our model trained with the \name data is robust develops deep video-language alignment understanding. We find that our model outperforms the baseline models on various experiments including video-language entailment (\S \ref{sec:exp_entailment}), NLE generation (\S \ref{sec:exp_nle}), and downstream tasks (\S \ref{sec:downstream}). 

\subsection{Performance on VideoCon Entailment Task}
\label{sec:exp_entailment}

We present the ROC-AUC scores of the tested models in Table \ref{tab:vc_entailment}. From the table we see that the baseline models find the \name testset difficult, as reflected by low AUC scores  (e.g. \preowl - $57.2$), close to random. Even training on \name train instances, but with ``easy’’ negatives (\randowl - $59.7$), hardly improves the base models. A significant improvement is achieved with the VNLI-specific model (67), showing that the entailment task is not inherently represented in generic video-language aligned training sets and requires specific training. Yet, the best performance is achieved by training on \name, which addresses the diversity in plausible misalignments and includes ``difficult’’ training examples, reaching $84.6$ AUC. This demonstrates the merit of \name for improving video-language alignment robustness. We show qualitative examples for the model predictions in \S \ref{sec:qualitative}.
%We compare the ROC-AUC score for the baseline models and our finetuned model in Table \ref{tab:vc_entailment}. We find that our model outperforms the video-language \mplugowl by a large margin $27$ points on the \name (LLM) test set. This indicates the our model can robustly evaluate the alignment between the raw video and the LLM-assisted contrast captions. Additionally, we find that \mplugowl ft. Random only leads to marginal gains on our dataset, thus highlighting at the usefulness of active collection of high-quality contrast captions instead of passively finetuning the model with additional data. 

When evaluating on out-of-domain (OOD) data around video types and misalignment distribution, we again see that training with \name offers significant improvement to alignment detection, outperforming all baselines, albeit with smaller relative gains: 17\% and 16\% improvement compared to \preowl on (Human) and (Human-Hard) respectively compared to 48\% on (LLM) test. In future work, we plan to further diversify the misalignments \name covers to further improve its benefits on OOD cases. 
%As we train our finetuned model on the \name training set, the increase in performance on its in-domain test set is expected. To assess the generalization capability of our model, we compare its ROC-AUC scores with the baseline models on the \name (Human) evaluation set (\S \ref{sec:data_gen_human}). We find that our model outperforms the atemporal model and video-language \mplugowl by $6$ and $12$ points, respectively. We further observe that the performance of our model also outperforms all the baseline models by achieving $74.4$ score on the \name (Human-Hard) evaluation set. Our results indicate that training with \name elicits robustness to a diverse range of contrast captions. We show qualitative examples for the model predictions in \S \ref{sec:qualitative}.

We notice that the performance of the VNLI atemporal model is better than existing video-language alignment  models. It might be attributed to its training with contrast captions in \cite{yarom2023you}. It further highlights that the existing video-language models are not robust in comparison to a atemporal probe on video-language alignment evaluation, corroborating the findings from \cite{buch2022revisiting,lei2022revealing}.
%Throughout our evaluation, we notice that the ROC-AUC of the atemporal model is better than the existing video-language models. It might be attributed to its training with contrast captions in \cite{yarom2023you}. It further highlights that the existing video-language models are not robust in comparison to a atemporal probe on video-language alignment evaluation, corroborating the findings from \cite{buch2022revisiting,lei2022revealing}.

\subsection{Performance on NLE Generation Task}
\label{sec:exp_nle}

%In practice, we can benefit from having natural language explanation (NLE) for fine-grained understanding of the difference between the video and the contrast caption. Like \cite{scheurer2023training} for large language models, our model's NLE can help in improving video-language modeling through feedback.

Table \ref{tab:vc_nle} presents the performance of the tested models against the ground-truth on the NLE task, depicting average $Q^2$ score and PaLM-2 entailment accuracy. The results show that on in-domain \name, \ftowl outperforms \preowl by an impressive 263\% and 178\% relative increase on $Q^2$ score and PaLM-2 accuracy respectively. This indicates the finetuned model can accurately generate NLE that match well with the ground-truth NLE. 
This indicates that our model can generate accurate NLE for a wide range of misalignments in the video captions, which makes it useful for dense video-language alignment evaluation.
%In Table \ref{tab:vc_nle}, we evaluate the generated Natural Language Explanations NLE against the ground-truth NLE using two automatic metrics: the text-based entailment score, ranging from $0$ to $1$, as determined by the $Q^2$ framework \cite{honovich2021q}, and the entailment accuracy, designated as $1$ for entailment and $0$ for contradiction, as computed by the PaLM-2 \cite{anil2023palm}. We find that our model outperforms the baseline model by average $Q^2$ entailment score of $0.31$ and average PaLM-2 entailment accuracy of $29\%$ on the \name (LLM) test set. This indicates the finetuned model can accurately generate NLE that match well with the ground-truth NLE.

On out-of-domain \name, the improvement is more moderate but still high: 40\% and 20\% relative increase on $Q^2$ and PaLM-2 respectively. This is probably due to the more diverse ways humans express explanations compared to LLM prompting. In future work we plan to further address linguistic diversity in explanations for more robust generation and evaluation.
%To assess the generalization of our model, we evaluate it against the human-written NLEs. We find that our model still outperforms the baseline model on the unseen video and NLE distribution by $0.1$ entailment score from $Q^2$ and $8\%$ PaLM-2 entailment accuracy. Finally, we find that the automatic methods achieve high agreement with human judgements, thus establishing their efficacy for scalable NLE generation evaluation (Appendix \S \ref{sec:human_agreement_nle}).

\subsection{Performance on Video-Text Downstream Tasks}
\label{sec:downstream}

%We aim to assess the impact of training a video-language model with the contrast captions on downstream tasks. Specifically, we evaluate the zero-shot text-to-video retrieval performance on the temporal-extensive SSv2-Temporal \cite{sevilla2021only} and action-centric SSv2-Events \cite{bagad2023test} (\S \ref{sec:t2vretrieval}). In addition, we assess the zero-shot performance on ATP-Hard \cite{buch2022revisiting} video question answering, that requires rich temporal and causal understanding of the events in the video, beyond object-centric static appearances (\S \ref{sec:vidqa}). In these experiments, we prompt the video-language generative model with the $M(V,Y)$ entailment template to calculate the alignment (Eq. \ref{eq:entailment}) between the raw video $V$ and the text $Y$. We provide more details about the downstream datasets and evaluation setup in Appendix \S \ref{sec:dets_downstream}.

\begin{figure*}[h]
    \centering
    \includegraphics[scale=0.5]{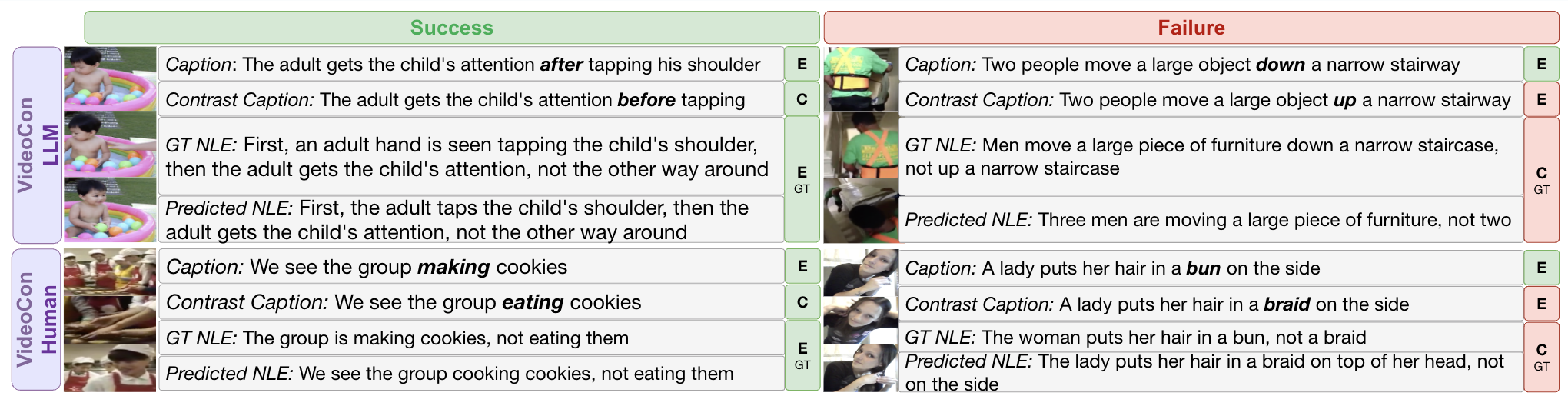}
    \caption{\small{\textbf{Qualitative examples for the success (green) and failure (red) modes of our model.} In every example, we present a few video frames in an temporal order from top to bottom, its associated caption, contrast caption, ground-truth NLE from the datasets. Additionally, we present the predicted NLE from our model. The small boxes at the end of caption cells indicate whether our model consider that caption to be grounded in the video. \textbf{E} and \textbf{C} indicates that the model predicts the caption to entail and contradict to the video, respectively. \textbf{E-GT} and \textbf{C-GT} indicates the predicted NLE entails and contradicts the ground-truth (GT) NLE, respectively.}}
    \label{fig:qualitative}
\end{figure*}

\begin{table}[h]
\centering
\resizebox{\linewidth}{!}{%
\begin{tabular}{lcccc}
\hline
                 \textbf{Models}      & \makecell{\textbf{SSv2-Temporal}\\\textbf{mAP}} &  \makecell{\textbf{SSv2-Events}\\\textbf{mAP}}     \\
                       \hline
Random                 & 7.3 & 3.3 \\
VideoCLIP & 9.8 & 6.4 \\
ImageBind (video-language) & 10.5 & 5.5\\
\preowl & 10.9 & 6.8 \\
TACT \cite{bagad2023test} & - & 7.8 \\ 
\randowl & 12.1 & 9.9 \\ 
\pali \cite{yarom2023you} & 14.6 & 10.4 \\
\ftowl (\textbf{Ours})          & \textbf{15.2}  & \textbf{11.4}\\\hline
\end{tabular}%
}
\caption{Mean Average Precision (mAP) scores for the tested models in the zero-shot text-to-video retrieval tasks.}
\label{tab:ssv2_zs}
\end{table}

%\subsubsection{Text-to-Video Retrieval}
%\label{sec:t2vretrieval}

We next present our results on the two downstream tasks, Text2Video Retrieval and Video Question Answering. Starting with the retrieval task, we report mean Average Precision (mAP) of the tested models on the SSv2-Temporal and SSv2-Events datasets in Table \ref{tab:ssv2_zs}. 
The benefits of training with additional examples tailored for temporal video-language alignment is already evident in the performance of \randowl, which improves over the previous SSv2-Events SOTA - TACT with a relative increase of 27\%.
%We compare the mean average precision (mAP) of the baseline models and our finetuned model on the SSv2-Temporal and SSv2-Events datasets in Table \ref{tab:ssv2_zs}. We consider the SSv2-Template captions instead of the label captions in our work since they are remove the object-centric bias in model evaluation \cite{lei2022revealing}. We find that our model achieves state-of-the-art results and outperforms the baseline \mplugowl model by $4.3$ mAP and $4.6$ mAP on the SSv2-Temporal and SSv2-Events, respectively. This indicates that our model has high temporal understanding which can be attributed to the presence of temporally-challenging instances in the \name dataset.

However, when training on harder negative contrastive instances, \ftowl achieves a significant improvement, outperforming all baselines, with a relative increase over the best baseline \pali model by 7.5\% on SSv2-Temporal and 9.6\% on SSv2-Events  (46\% over TACT), setting new SOTA results. This points at the benefits of exposing the model to temporal examples, such as \emph{actions} and \emph{event-order}.
%In addition, we observe that our model beats the previous state-of-the-art TACT on the SSv2-Events convincingly. This indicates that our model can understand the alignment for the videos depicting multi-events accurately. This can be attributed to the temporal and event-order understanding developed as part of learning to distinguish `event order flipping' misalignment in the \name dataset. 

%\subsubsection{Video Question Answering}
%\label{sec:vidqa}

\begin{table}[h]
\centering
\begin{tabular}{lc}
\hline
\textbf{Models}          & \textbf{Accuracy (\%)} \\\hline
CLIP           & 23.8     \\
VideoCLIP            & 23.4     \\
ImageBind (video-language)      & 25.4     \\
TACT \cite{bagad2023test} & 27.6 \\
VFC \cite{momeni2023verbs}           & 31.4     \\
\preowl     & 37.1     \\
\randowl & 37.2 \\
\pali \cite{yarom2023you} & 39.0 \\
\ftowl (\textbf{Ours})  & \textbf{41.1}\\\hline
\end{tabular}
\caption{
Accuracy scores for the tested models on the zero-shot video question-answering task on ATP-Hard dataset.
%Comparison between the baseline models and our model on the zero-shot text-to-video video question-answering task on ATP-Hard dataset. We find that our model outperforms the existing models by a large margin on the dataset, suggesting at its high causal and temporal video-language understanding.
}
\label{tab:vqa_atphard}
\end{table}

For the Video Question Answering task, we compare the performance of the various models in Table \ref{tab:vqa_atphard}. Here too \ftowl achieves SOTA results and outperforms the strongest baseline \pali model with a relative increase of 5.1\%.
This corroborates the observations in our other experiments, which demonstrate the advantage of the \name datasets, covering various misalignments, especially those pertaining to temporal and causal reasoning  over dynamic events. The results also confirm the need for carefully chosen contrastive negative examples, showing that picking negatives at random may mask out the potential benefit of an alignment training set. Finally, the competitive performance of atemporal \pali model on the downstream tasks is surprising and underscores the need for stronger video-language datasets for robust benchmarking.

%Since our model performs robust video-language alignment evaluation, we aim to assess its performance on the video question answering (QA) task \cite{xiao2021next} due to its practical importance. We first convert the question-answer pairs in the temporal-causal video QA dataset \cite{buch2022revisiting} into candidate video captions using LLM (Appendix \ref{appen_sec:videoqa}). Subsequently, we assess the entailment score of the candidate captions for their grounding in the video.
%Finally, we compare the performance of the baseline models and our finetuned model in Table \ref{tab:vqa_atphard}. We find that our model achieves state-of-the-art results and outperforms the baseline \mplugowl model by $4\%$ accuracy on the dataset.

%We further observe that our model outperforms the action-aware VFC \cite{momeni2023verbs} and time-aware TACT model. This indicates that our model can perform temporal and causal reasoning over the static objects and dynamic events grounded in the ATP-Hard dataset videos.  

\begin{figure}[h]
    \centering
    \includegraphics[scale=0.35]{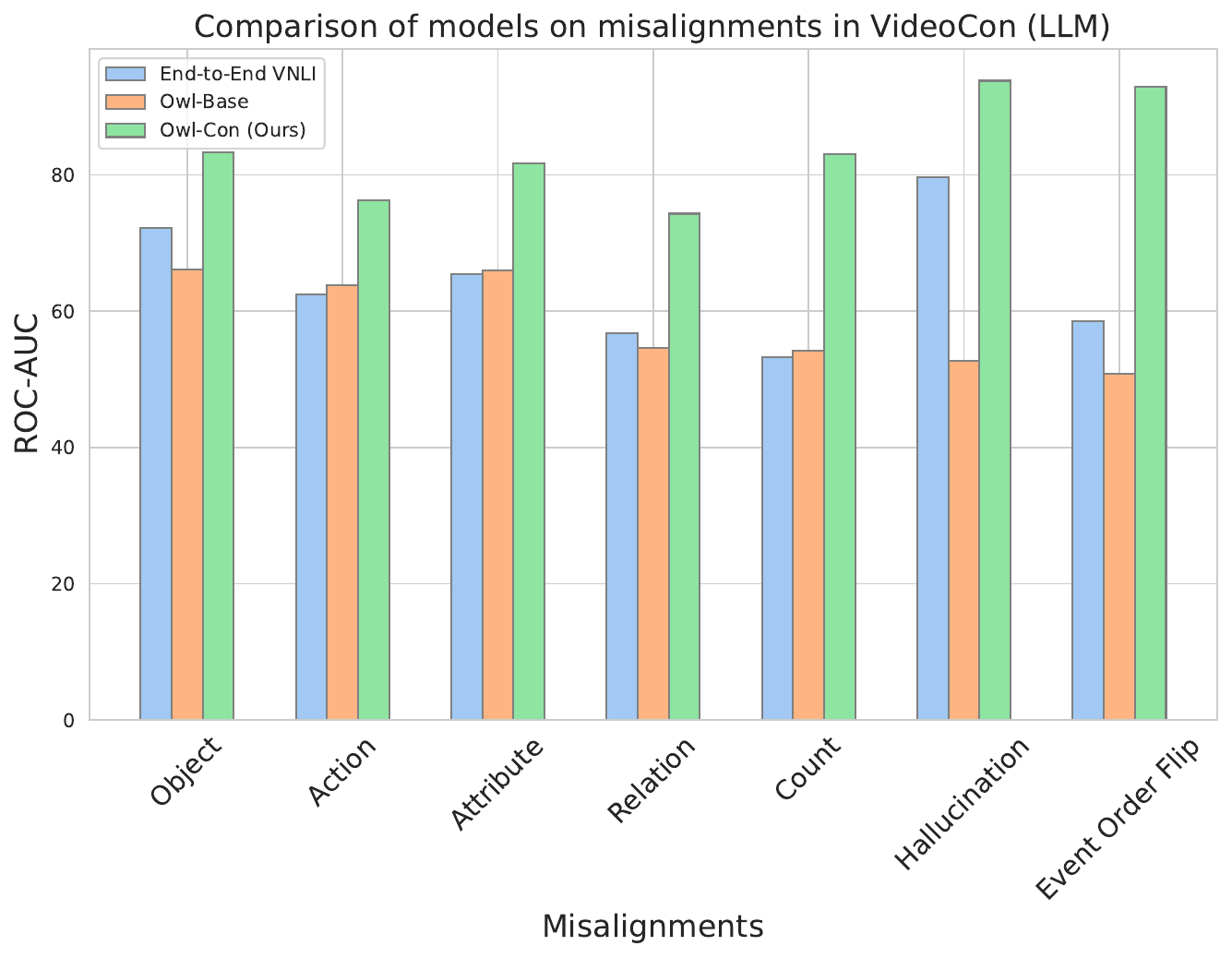}
    \caption{\small{ROC-AUC of \pali, \preowl, and \ftowl across all types of misalignment in \name (LLM) test set.}}
    \label{fig:per_misalignment}
\end{figure}

\section{Analysis}
\label{sec:analysis}

We analyze \ftowl’s performance improvements across the kinds of misalignments in \name. Additionally, we present a few qualitative examples to highlight the success and failure modes of our model.
%We study the performance improvements across the kinds of misalignments in \name dataset (\S \ref{sec:per_misalignment_entailment}). Finally, we discuss a few qualitative examples to highlight the success and failure modes of our model (\S \ref{sec:qualitative}).

\subsection{Per-misalignment Entailment Results}
\label{sec:per_misalignment_entailment}

We compared the ROC-AUC scores of the atemporal \pali, \preowl, and \ftowl on specific misalignments in the contrast captions from \name (LLM) testset in Figure \ref{fig:per_misalignment}. We observed that \ftowl outperforms the baseline models across all misalignment types. This suggests that our model can reason well about the entities, their relations, and the temporal order of events in the video. 
%We compare the ROC-AUC score of the atemporal (\pali), pretraining \mplugowl, and our finetuned model on specific misalignments in the contrast captions from \name (LLM) dataset in Figure \ref{fig:per_misalignment}. We observe that our finetuned model outperforms the baseline models across all the type of misalignments. This suggests that the model can reason well about the entities, their relations, and the temporal order of events grounded in the video. We attribute the improvements on the unseen \name (Human-Hard), and downstream tasks to these robust alignment evaluation capabilities of our model.

% \yb{candidate to write shorter/remove}Specifically, we find that the ROC-AUC score is highest for the \textit{hallucination} and \textit{event order flip}. In addition, we find that the atemporal struggles at \textit{counting} misalignment due to the lack of such instances in its training data, and \textit{event order flip} due to highly temporal nature of the misalignment. Finally, we observe that the performance improvements on specific misalignments such as \textit{action} and \textit{relation} are lower than the other types , which can be mitigated by more diverse data or improved training paradigms.  

The largest improvement of \ftowl compared to the two baselines is on \emph{event order flip}, indicating that the baselines lack temporal understanding and the \name is efficient in adding this capability to an alignment model. 
In addition, on \emph{hallucination} both \ftowl and \pali significantly outperform \preowl, since both models were explicitly exposed to entailment/non-entailment training data. 
It is surprising to see that while \pali was trained on significantly more entailment data, much of it human-curated, \ftowl outperforms it with only automatically generated data. This could be due to the better encoding of video in \ftowl compared to the atemporal nature of \pali.
Finally, the analysis shows other types of atemporal misalignments that are difficult for \pali to sort out,  e.g. \emph{counting'} and \emph{relation}, where the training data in \name is useful to improve these capabilities as well. This shows that our approach of detailed analysis of misalignment types of generation of examples for them is effective.
%Our findings indicate that our model achieves the highest ROC-AUC scores for `hallucination' and `event order flip' misalignments. The performance on `action' and `relation' misalignments is lower, suggesting that either more varied data could help. In addition, the \pali model underperforms in identifying `counting' errors, likely due to the lack of such instances in its training data, and struggles with `event order flip' as it is atemporal. 

\subsection{Qualitative Examples}
\label{sec:qualitative}

We highlight a few classification examples of \ftowl in Figure \ref{fig:qualitative}. The rows refer to the test  source of the instances and the columns refer to the success and failure modes, respectively. In Row1/Column1, we observe that our model provides correct predictions for the entailment between the video and original caption while predicting contradiction for the contrast caption that flips the order of the events i.e., \textit{grabbing attention} and \textit{tapping shoulders}. Interestingly, our model can also provide the accurate NLE when prompted with the video and the contrast caption. This suggests that our model is useful for providing fine-grained details about the video-language alignment. In Row2/Column2, the model confuses `buns' with `braids' in hair and gives a wrong NLE that contradicts the ground-truth. This error, due to its inability to distinguish between objects, might be improved by expanding the variety and contrast in the dataset's videos and captions.
%We highlight a few qualitative examples to highlight in Figure \ref{fig:qualitative}. The rows of the figure represent the source of the instances and the columns represent the success and failure modes, respectively. In Row1-Column1, we observe that our model provides correct predictions for the entailment between the video and original caption while predicting contradiction for the contrast caption that flips the order of the events i.e., \textit{grabbing attention} and \textit{tapping shoulders}. Interestingly, our model can also provide the accurate NLE when prompted with the video and the contrast caption. This suggests that our model is useful for providing fine-grained details about the video-language alignment. In Row2-Column2, the model confuses `buns' with `braids' in hair and gives a wrong NLE that contradicts the ground-truth. This error, due to its inability to distinguish between objects, might be improved by expanding the variety and contrast in the dataset's videos and captions.

% \subsection{Ablation Studies}

% 1. Training with just the entailment prompt instead of the mix of the entailment and natural language prompt
% 2. How does the performance look when the model are not trained with the event order misalignment?
\section{Related Work}
\label{sec:related_work}

\paragraph{Foundation Models for Video-Language Understanding.} Foundation models have emerged for video-language understanding \cite{xu2021videoclip,wang2022internvideo,arnab2021vivit,xu2023mplug,alayrac2022flamingo} by pre-training on large amount of video-text pairs scraped from the web \cite{xue2022hdvila,miech2019howto100m,bain2021frozen}. Additionally, prior works have either leveraged the pretrained CLIP model for video-language tasks \cite{luo2022clip4clip,fang2021clip2video,ma2022x} or adopted a socratic approach \cite{zeng2022socratic,wang2022language} to employ LLMs (GPT-3) in reasoning over video captions. We highlight that despite the large-scale training of the video-language foundation models \cite{girdhar2023imagebind,xu2021videoclip,xu2023mplug}, they lack robustness to semantic changes to the captions (e.g., changing the temporal order of the events) which severely limits their real-world use for alignment applications. We provide a fix to the issue by training models on a novel video-centric \name dataset.

\paragraph{Improving Video-Language Robustness.}  Prior work \cite{park2022exposing,momeni2023verbs,wang2023paxion} highlights that the video-text models cannot comprehend the semantics of the text with focus on manipulating the verb, actions, and entities grounded in the video description. To improve the temporal understanding, \cite{bagad2023test} finetunes a pretrained model with temporal order loss. Despite this, their models do not achieve good zero-shot performance on downstream tasks consistently and is highly dependent on the choice of the finetuning dataset. In our work, we categorize a wide range of plausible misalignments in the contrast captions, and create a temporally-challenging \name dataset. We show that \name enables robust training of the model that achieve state-of-the-art zero-shot performance on various video-language tasks.

\paragraph{Video-Language Alignment Evaluation.} Many applications such as text-to-video retrieval \cite{xu2016msr,Wang_2019_ICCV,goyal2017something} and text-to-video generation \cite{blattmann2023align,wang2023modelscope} require evaluation of the semantic alignment between the natural language text and raw video. In this work, we indicate that the existing video-text models such as VideoCLIP and ImageBind are not robust to semantic changes in the video captions, which becomes critical for faithful video-text alignment evaluation. Beyond this, prior work \cite{liang2020alice,scheurer2023training} has shown that fine-grained feedback can be useful for evaluating and training better models. In our work, we propose \name and finetune a video-language generative model to perform robust entailment task and provide fine-grained NLE for the observed misalignments between the video and text. In the future, our model can be utilized to enhance alignment through sparse (entailment scores) and dense (fine-grained NLE) feedback \cite{scheurer2023training}.
\section{Conclusion}
\label{sec:conclusion}
% \yb{Conclusion can be shorter. Consider making a separate "Limitations" section - it's also sometimes required (check CVPR instructions)}
% We presented \name, a temporally-extensive dataset for robust video-text alignment. The main benefits of \name stem from including a wide range of semantically plausible misalignments within contrast captions and the natural language explanation (NLE) for the discrepancy between the grounding of the text in the video. Empirically, we show that finetuning a generative video-language model for the video-text entailment task and NLE generation task improves its robustness to unseen (human-constructed) instances. Finally, we show that our finetuned model achieves state-of-the-art performance on temporally-challenging downstream tasks including text-to-video retrieval and video question answering. 

We introduced a comprehensive dataset, \name, designed for robust video-text alignment. It features various semantic misalignments and explanations for text-video discrepancies. Through finetuning video-language models on this dataset, we enhanced their performance on complex tasks like text-to-video retrieval and video question answering, achieving state-of-the-art results.

One current limitation and an important future direction is to increase the complexity of the generated contrast captions. Specifically, the model may encounter several misalignments within a single contrast caption. Addressing this issue, the model should be equipped to accurately assign low entailment scores to these contrast captions and consequently generate precise NLEs. An important future direction is to scale \name to larger datasets. Here, we create contrast captions for high-quality captions written by humans for every video, however, the web-scale datasets have low-quality captions that are not well grounded in the video. In this regard, using synthetic data followed by \name-like contrast caption generation can be a plausible approach \cite{nguyen2023improving}. Further, it would be important to scale our \name (Human) dataset more comprehensively to cover a larger set of visual domains (e.g., generated videos), contrast captions and NLE for robust evaluation.

{
    \small
    \bibliographystyle{ieeenat_fullname}
    \bibliography{main}
}

\clearpage
\setcounter{page}{1}
\maketitlesupplementary
\appendix
\section{Detailed Related Work}
\label{sec:detailed_related_work}

\paragraph{Foundation Models for Video-Language Understanding.} Towards the goal of building general-purpose AI systems, instantiations such as GPT-3 \cite{brown2020language}, CLIP \cite{xu2021videoclip}, ALIGN \cite{jia2021scaling} have scaled up self-supervision within single modality (e.g., text) or multiple modalities (e.g., vision-language) by utilizing vast amount of data from the web \cite{CommonCrawl,schuhmann2022laion}. Post-training, these models can solve a wide range of downstream tasks through few-shot learning or task-specific finetuning. Similar foundation models have emerged for video-language understanding \cite{xu2021videoclip,wang2022internvideo,arnab2021vivit,xu2023mplug,alayrac2022flamingo} by pre-training on large amount of video-text pairs scraped from the web \cite{xue2022hdvila,miech2019howto100m,bain2021frozen}. In addition, prior works have either leveraged the pretrained CLIP model for video-language tasks \cite{luo2022clip4clip,fang2021clip2video,ma2022x} or adopted a socratic approach \cite{zeng2022socratic,wang2022language} to employ LLMs (GPT-3) in reasoning over video captions. We highlight that despite the large-scale training of the video-language foundation models \cite{girdhar2023imagebind,xu2021videoclip,xu2023mplug}, they lack robustness to semantically plausible contrast captions (e.g., changing the temporal order of the events) which severely limits their real-world use for alignment applications. We provide a fix to the issue by creating a novel video-centric \name dataset for robust training.

\paragraph{Improving Video-Language Robustness.}  Prior work \cite{park2022exposing,momeni2023verbs,wang2023paxion} highlights that the video-text models cannot comprehend the semantics of the text with focus on manipulating the verb and entities grounded in the video description. At the same time, \cite{bagad2023test,wang2023paxion} indicate that the video-text models are not robust to the temporal order of events depicted in the video. To improve the temporal understanding, \cite{bagad2023test} finetunes a pretrained model with temporal order loss. Despite this, their models do not achieve good zero-shot performance on downstream tasks consistently and is highly dependent on the choice of the finetuning dataset. In our work, we categorize a wide range of plausible misalignments in the contrast captions, 7 in total, and create a temporally-challenging \name dataset by filtering image-temporally-easy instances using a image-text alignment model. Our dataset also covers a wide range of video-text domains covered in \msr, \vatex, and \tempo datasets. Finally, we show that \name enables robust training of the model that achieve state-of-the-art zero-shot performance on various video-language tasks.

\paragraph{Video-Language Alignment Evaluation.} Many traditional applications such as text-to-video retrieval \cite{xu2016msr,Wang_2019_ICCV,goyal2017something} require evaluation of the semantic alignment between the natural language text and raw video. With the rise of creative generative models \cite{ramesh2021zero,rombach2022high}, recent methods \cite{hu2023tifa, yarom2023you} have emerged for robust and faithful evaluation of the alignment between the input text and generated image. Similarly, we would soon require robust video-language alignment evaluation to assess the faithfulness of upcoming text-to-video generative models \cite{blattmann2023align,wang2023modelscope}. In this work, we indicate that the existing video-text models such as VideoCLIP and ImageBind are not robust to semantic changes in the video captions, which becomes critical for faithful video-text alignment evaluation. Beyond this, prior work \cite{liang2020alice,scheurer2023training} has shown that fine-grained feedback can be useful for evaluating and training better models. In our work, we propose \name and finetune a video-language generative model to perform robust entailment task and provide fine-grained natural language explanations for the observed misalignments between the video and text. As a result, we achieve large performance gains on unseen \name (Human) test set as well as downstream tasks.  

\section{Details about Video-Language Datasets}
\label{sec:details_vld}

\paragraph{\msr} \cite{xu2016msr} is a large-scale video descriptions dataset covering a wide range of daily life categories ranging from music to cooking. Originally, the dataset contains $10$K videos with $20$ human-written descriptions for every video. The duration of the video clips in the dataset is between $10$-$30$ seconds. In our work, we filter the videos that are no longer publicly available on Youtube. As a result, we removed $29\%$ of the videos. We utilize the video-text data from \msr train-val set for \name train-val set, and \msr test set for \name test set. 

\paragraph{\vatex} \cite{Wang_2019_ICCV} is large-scale dataset that is focused on enhanced the linguistic complexity and diversity of the video descriptions. The dataset consists of $600$ human activities video content from the Kinetics-600 \cite{kay2017kinetics}. Originally, the dataset contains $26$K videos in the train set and $3$K videos in the validation set with $10$ human-written descriptions for every video. We used half of the \vatex training set for \name train-val set and half of the \vatex validation set for \name test set. Further, we filter the videos that are no longer publicly available on Youtube. As a result, we removed $23\%$ of the videos.

\begin{figure}[h]
    \centering
    \includegraphics[scale=0.7]{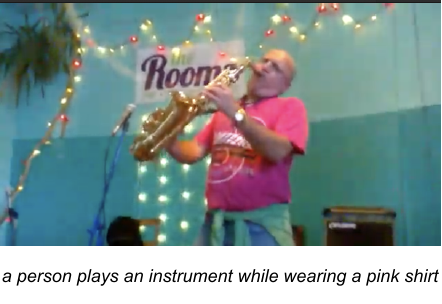}
    \caption{\small{\textbf{Illustration of a temporally-easy instance (video-text pair) from the \vatex dataset.} We observe that the video caption (`a person ... pink shirt') is well-grounded in just a single frame of the video. As a result, the video-text models are not incentivized to develop video-centric understanding (e.g., temporality) while training on such instances.}}
    \label{fig:temp_easy_example}
\end{figure}

Since \msr and \vatex are general-purpose datasets collected from the web, prior work \cite{buch2022revisiting,lei2022revealing} has shown that many of the video-text pairs in these datasets are not temporally-challenging. As shown in Figure \ref{fig:temp_easy_example}, a single frame from a \vatex dataset video shares sufficient semantic information with the video caption, and hence it is not temporally-challenging. The abundance of such instances in the dataset do not encourage the models to develop robust video-language understanding capabilities. Hence, we utilize \pali model \cite{yarom2023you} to filter temporally-easy instances and make \name temporally-extensive.

\paragraph{\tempo} \cite{hendricks2018localizing} is an unique temporal reasoning video-text dataset. The dataset is constructed from merging two 5 second segments of the videos in the DiDeMo dataset \cite{anne2017localizing}. \tempo dataset consists of two versions -- template-based (TL) and human-written (HL). In our work, we use the video-captions from the \tempo (HL) dataset. The \name consists of $11$K \tempo training video-text pairs for its train-val set, and $1.8$K \tempo testing video-text pairs for its testing set.

Overall, \name has $27$K and $5$K unique videos for training-validation and testing, respectively. In addition, it consists $62$K and $13$K unique captions for training-validation and testing, respectively.

\section{Misalignment Assignment}
\label{sec:misalignment_assignment}

Here, we assign the type of misalignment within the contrast caption for a given video caption. The video caption and the assigned misalignment is then used to prompt large language model (LLM) to generate the contrast caption. 

We consider instances from the datasets $(V, T)$ where $V$ is the video caption and $T$ is the text caption. If the caption contains one of the keywords from Table \ref{tab:relation_map}, we assign \textit{relation} misalignment to it.  If the caption contains a number (`one' - `ten'), we assign \textit{count} misalignment to it.

\begin{table}[h]
\begin{tabular}{|p{8cm}|}
\hline
 `above', `below', behind', `in front of', `top of', `under', `inside', `outside', `beneath', `left of', `right of', `upwards', `downwards', `up', `down', `far away', `towards'  \\\hline                     
\end{tabular}
\caption{The list of keywords that indicate spatial relations between entities in the video captions.}
\label{tab:relation_map}
\end{table}

For the instances from \tempo dataset, the captions are assigned \textit{object}, \textit{action}, \textit{attribute}, \textit{hallucination}, \textit{event order flipping} misalignments with equal probability. For the instances from the \msr and \vatex dataset, we identify whether the $(V,T)$ instance is temporally-easy $(V, T)_{\text{easy}}$ or temporally-challenging $(V, T)_{\text{challenging}}$ using the \pali model, as described in \S \ref{sec:temp_challenging_vl_data}. For the temporally-challenging instances $(V, T)_{\text{challenging}}$, we utilize the PaLM-2 LLM API to identify whether the video caption $T$ describes multiple events $Ev$. For example, `a girl walks down a hill and eats icecream' has two events i.e., `walking down a hill' and `eating icecream' ($Ev = \text{multiple}$). On the other hand, `a person moving a toy away from the child' consists only a single event ($Ev = \text{single}$). We assign \textit{event order flipping} misalignment to all the captions from $(V, T)_{\text{challenging}}$. We assign \textit{object}, \textit{action}, \textit{attribute}, and \textit{hallucination} misalignment with equal probability to the captions from $(V, T)_{\text{easy}}$. 

We use Spacy \cite{spacy2} to extract POS tags for the words in the video caption. We ensure that the captions without any adjective, verb, noun parts-of-speech words in the captions are not assigned \textit{attribute}, \textit{verb}, and \textit{object} misalignment, respectively.

\section{LLM Prompt}
\label{sec:llm_prompt}

We present the prompts used to generate contrast captions for \name dataset in Figure \ref{tab:prompt_object} - \ref{tab:prompt_event}. We have separate prompts for every misalignment where we provide the task description, guidelines, and a few in-context examples. In our work, we use PaLM-2 LLM API. Specifically, we utilize `chat-bison@001' with chat parameters temperature = $0.5$, max output tokens = $256$, top p = $0.95$, and top k = $40$.

\begin{figure*}
\centering
\resizebox{\linewidth}{!}{

\begin{tabular}{p{1.3\linewidth}}

\toprule
Your objective is to generate a contradiction sentence using a provided ``input sentence" based on a specific ``misalignment scenario" called ``Object Misalignment''. In this scenario, you should modify a key object in the "input sentence".\\\\

Please also identify the portion of the ``input sentence" you've expanded and label this as the ``source." Then, specify the new elements introduced in the ``sentence + object misalignment" as the ``target".\\\\

Your last task is to provide a ``Correct Misalignment" description, clarifying how the ``input sentence" is different from the ``sentence + object misalignment".\\\\

Key Requirements:
- The ``sentence + object misalignment" should be plausible and could theoretically occur in real life. \\\\

Guidelines:\\
1. The ``sentence + object misalignment" should be clearly distinguishable from the ``input sentence".\\
2. Your replacements should be creative yet reasonable.\\
3. Avoid changing gender, color, or race of humans in the sentence.\\
4. The ``Correct Misalignment" should describe how the "input sentence" diverges from the ``sentence + object misalignment".\\\\

Input Sentence: a smartphone and a finger pointing to the bluetooth buttons\\
Sentence + Object Misalignment: a smartphone and a toe pointing to the bluetooth buttons\\
Source: ``finger"\\
Target: ``toe"\\
Correct Misalignment: a finger is pointing to the bluetooth buttons instead of a toe\\\\

Input Sentence: woman plays a song on the piano\\
Sentence + Object Misalignment: woman plays a song on the cello\\
Source: ``piano''\\
Target: ``cello"\\
Correct Misalignment: woman plays a song on the piano instead of cello\\\\

Input Sentence: a man is going in the wheel skate\\
Sentence + Object Misalignment: a man is going in the bicycle\\
Source: ``wheel skate"\\
Target: ``bicycle"\\
Correct Misalignment: a man is going in the wheel skate instead of the bicycle\\\\

Now it's your turn.\\\\

Input Sentence: $<$insert caption$>$\\
Sentence + Object Misalignment:\\
Source:\\
Target:\\
Correct Misalignment:\\
\bottomrule
\end{tabular} }

\caption{PaLM-2 LLM API prompt to generate contrast captions with \textit{Object} misalignment.}
\label{tab:prompt_object}
\end{figure*}

\begin{figure*}
\centering
\resizebox{\linewidth}{!}{

\begin{tabular}{p{1.3\linewidth}}
\toprule

Your objective is to generate a contradiction sentence using a provided ``input sentence" based on a specific ``misalignment scenario" called ``Action Misalignment." In this scenario, you should modify specific action performed by the object in the ``input sentence".\\\\

Please also identify the portion of the ``input sentence" you've expanded and label this as the ``source". Then, specify the new elements introduced in the "sentence + action misalignment" as the ``target".\\\\

Your last task is to provide a ``Correct Misalignment" description, clarifying how the ``input sentence" is different from the ``sentence + action misalignment".\\\\

Key Requirements:\\
- The ``sentence + action misalignment" should be plausible and could theoretically occur in real life.\\\\

Guidelines:\\
1. The ``sentence + action misalignment" should be clearly distinguishable from the ``input sentence".\\
2. Your replacements should be creative yet reasonable.\\
3. Avoid changing gender, color, or race of humans in the sentence.\\
4. The ``Correct Misalignment" should describe how the "input sentence" diverges from the ``sentence + action misalignment".\\\\

Input Sentence: a person repairing the car\\
Sentence + Action Misalignment: a person driving the car\\
Source: ``repairing"\\
Target: "driving"\\
Correct Misalignment: a person is repairing the car instead of the driving it\\\\

Input Sentence: a woman is singing\\
Sentence + Action Misalignment: a woman is yelling\\
Source: ``singing"\\
Target: ``yelling"\\
Correct Misalignment: a woman is singing instead of yelling\\\\

Input Sentence: an animated cartoon of a monster catching a man by the foot and then launching him like a slingshot\\
Sentence + Action Misalignment: an animated cartoon of a monster throwing a man by the foot and then launching him like a slingshot\\
Source: ``catching a man"\\
Target: ``throwing a man"\\
Correct Misalignment: a monster is catching a man instead of throwing a man\\\\

Input Sentence: a robot is entering a hall talking to a person\\
Sentence + Action Misalignment: a robot is leaving a hall talking to a person\\
Source: ``entering"\\
Target: ``leaving"\\
Correct Misalignment: a robot is entering a hall not leaving it\\\\

Now it's your turn.\\\\

Input Sentence: $<$insert caption$>$ \\
Sentence + Action Misalignment:\\
Source:\\
Target:\\
Correct Misalignment:\\

\bottomrule
\end{tabular} }

\caption{PaLM-2 LLM API prompt to generate contrast captions with \textit{Action} misalignment.}
\label{tab:prompt_action}
\end{figure*}

\begin{figure*}
\centering
\resizebox{\linewidth}{!}{

\begin{tabular}{p{1.3\linewidth}}
\toprule
Your objective is to generate a contradiction sentence using a provided ``input sentence" based on a specific ``misalignment scenario" called "Counting Misalignment". In this scenario, you should modify the mathematical count of the objects in the ``input sentence".\\\\

Please also identify the portion of the ``input sentence" you've expanded and label this as the "source". Then, specify the new elements introduced in the ``sentence + counting misalignment" as the ``target".\\\\

Your last task is to provide a ``Correct Misalignment" description, clarifying how the ``input sentence" is different from the ``sentence + counting misalignment".\\\\

Key Requirements:\\
- The ``sentence + counting misalignment" should be plausible and could theoretically occur in real life.\\
- Only focus on the counts of the objects; do not replace or remove any existing objects, actions or attributes in the "input sentence."\\\\

Guidelines:\\
1. The ``sentence + counting misalignment" should be clearly distinguishable from the ``input sentence".\\
2. Avoid changing gender, color, or race of humans in the sentence.\\
3. The ``Correct Misalignment" should describe how the ``input sentence" diverges from the ``sentence + counting misalignment".\\\\

Input Sentence: a man is entering a room with three surgeons\\
Sentence + Counting Misalignment: a man is entering a room with one surgeon\\
Source: ``three surgeons"\\
Target: ``one surgeon"\\
Correct Misalignment: the man enters the room with three surgeons instead of one surgeon\\\\

Input Sentence: three girls singing on stage on the voice\\
Sentence + Counting Misalignment: six girls singing on stage on the voice\\
Source: ``three girls"\\
Target: ``six girls"\\
Correct Misalignment: three girls are singing on the voice instead of six girls\\\\

Input Sentence: a video showcasing 6 different peoples reactions to a certain video the video seemed family oriented\\
Sentence + Counting Misalignment: a video showcasing 2 different peoples reactions to a certain video the video seemed family oriented\\
Source: ``6 different peoples reactions"\\
Target: ``4 different peoples reactions"\\
Correct Misalignment: six different people were showcasing their reactions to a video instead of four different people\\\\

Now it's your turn.\\\\

Input Sentence: $<$insert caption$>$\\
Sentence + Counting Misalignment:\\
Source:\\
Target:\\
Correct Misalignment:\\
\bottomrule
\end{tabular} }

\caption{PaLM-2 LLM API prompt to generate contrast captions with \textit{Count} misalignment.}
\label{tab:prompt_counting}
\end{figure*}

\begin{figure*}
\centering
\resizebox{\linewidth}{!}{

\begin{tabular}{p{1.3\linewidth}}
\toprule
Your objective is to generate a contradiction sentence using a provided ``input sentence" based on a specific ``misalignment scenario" called "Attribute Misalignment". In this scenario, you should modify an attribute of an object in the ``input sentence".\\\\

Please also identify the portion of the ``input sentence" you've expanded and label this as the ``source." Then, specify the new elements introduced in the ``sentence + attribute misalignment" as the ``target".\\\\

Your last task is to provide a ``Correct Misalignment" description, clarifying how the ``input sentence" is different from the ``sentence + attribute misalignment".\\\\

Key Requirements:\\
- The ``sentence + attribute misalignment" should be plausible and could theoretically occur in real life.\\\\

Guidelines:\\
1. The ``sentence + attribute misalignment" should be clearly distinguishable from the ``input sentence."\\
2. Your replacements should be creative yet reasonable.\\
3. Avoid changing gender, color, or race of humans in the sentence.\\
4. The ``Correct Misalignment" should describe how the ``input sentence" diverges from the ``sentence + attribute misalignment".\\\\

Input Sentence: man in blue shirt is test driving his new car\\
Sentence + Attribute Misalignment: man in red shirt is test driving his new car\\
Source: ``blue"\\
Target: ``red"\\
Correct Misalignment: a man in blue shirt instead of the red shirt\\\\

Input Sentence: a group of people playing with giant beach balls\\
Sentence + Attribute Misalignment: a group of people playing with small beach balls\\
Source: ``giant"\\
Target: ``small"\\
Correct Misalignment: a group of people playing with giant beach balls instead of the small beach balls\\\\

Input Sentence: there is a man with serious face looking cruelly\\
Sentence + Attribute Misalignment: there is a man with happy face looking kindly\\
Source: ``serious face looking cruelly"\\
Target: ``happy face looking kindly"\\
Correct Misalignment: a man is with the serious face looking cruelly instead of the happy face looking kindly\\\\

Now it's your turn.\\\\

Input Sentence: $<$insert caption $>$ \\
Sentence + Attribute Misalignment:\\
Source:\\
Target:\\
Correct Misalignment:\\
\bottomrule
\end{tabular} }
\caption{PaLM-2 LLM API prompt to generate contrast captions with \textit{Attribute} misalignment.}
\label{tab:prompt_attribute}
\end{figure*}

\begin{figure*}
\centering
\resizebox{\linewidth}{!}{

\begin{tabular}{p{1.3\linewidth}}
\toprule

Your objective is to generate a contradiction sentence using a provided ``input sentence" based on a specific ``misalignment scenario" called ``Relation Misalignment". In this scenario, you should change the relation between the objects in the sentence.\\\\

Please also identify the portion of the ``input sentence" you've expanded and label this as the ``source". Then, specify the new elements introduced in the ``sentence + relation misalignment" as the ``target".\\\\

Your last task is to provide a ``Correct Misalignment" description, clarifying how the ``input sentence" is different from the ``sentence + relation misalignment".\\\\

Key Requirements:\\
- The ``sentence + relation misalignment" should be plausible and could theoretically occur in real life.\\
- Relation is a word or group of words used before a noun, pronoun, or noun phrase to show direction, time, place, location, spatial relationships, or to introduce an object. Examples include: ``above", ``below", ``inside", ``outside", ``front of", ``behind", ``up", ``down", ``left", ``right" etc.\\
- Only focus on the relations between the objects; do not replace or remove any existing objects, actions or attributes in the ``input sentence".\\\\

Guidelines:\\
1. The ``target" should introduce a contradiction when compared to the "source," without being a mere negation.\\
2. The ``sentence + relation misalignment" should be clearly distinguishable from the ``input sentence".\\
3. Your additions should be creative yet reasonable.\\
4. Avoid changing gender, color, or race of humans in the sentence.\\
5. The ``Correct Misalignment" should describe how the ``input sentence" diverges from the ``sentence + relation misalignment".\\\\

Input Sentence: people are dancing and singing outside\\
Sentence + Relation Misalignment: people are dancing and singing inside the club\\
Source: ``outside"\\
Target: ``inside the club"\\
Correct Misalignment: people are dancing and singing outside, not inside the club\\\\

Input Sentence: a woman talking in front of a camera\\
Sentence + Relation Misalignment: a woman is talking behind a camera\\
Source: ``in front of a camera"\\
Target: ``behind a camera"\\
Correct Misalignment: a woman talks in front of a camera, not behind it\\\\

Input Sentence: a bowl of grey shrimp is shown above a yellow broth\\
Sentence + Relation Misalignment: a bowl of grey shrimp is shown below a yellow broth\\
Source: ``above"\\
Target: ``below"\\
Correct Misalignment: a bowl of grey shrimp is shown above a yellow broth, not below it\\\\

Input Sentence: a kid flips over a mattress on a trampoline\\
Sentence + Relation Misalignment: a kid flips over a mattress under the trampoline\\
Source: ``on a trampoline"\\
Target: ``under the trampoline"\\
Correct Misalignment: a kid flips the mattress on a trampoline, not under it\\\\

Input Sentence: the objects are placed far away from each other\\
Sentence + Relation Misalignment: the objects are placed close to each other\\
Source: ``far away"\\
Target: ``close"\\
Correct Misalignment: the objects are placed far away from each other, instead of close to each other\\\\

Now it's your turn.\\\\

Input Sentence: $<$insert caption$>$\\
Sentence + Relation Misalignment:\\
Source:\\
Target:\\
Correct Misalignment:\\
\bottomrule
\end{tabular} }
\caption{PaLM-2 LLM API prompt to generate contrast captions with \textit{Relation} misalignment.}
\label{tab:prompt_relation}
\end{figure*}

\begin{figure*}
\centering
\resizebox{\linewidth}{!}{

\begin{tabular}{p{1.3\linewidth}}
\toprule
Your objective is to generate a contradiction sentence using a provided ``input sentence" based on a specific ``misalignment scenario" called "Hallucination Misalignment". In this scenario, you should add new elements to the sentence without replacing or removing anything that is already there.\\\\

Please also identify the portion of the ``input sentence" you've expanded and label this as the ``source". Then, specify the new elements introduced in the ``sentence + hallucination" as the ``target".\\\\

Your last task is to provide a ``Correct Misalignment" description, clarifying how the ``input sentence" is different from the ``sentence + hallucination".\\\\

Key Requirements:\\
- The ``sentence + hallucination" should be plausible and could theoretically occur in real life.\\
- Only add elements; do not replace or remove any existing elements in the ``input sentence".\\\\

Guidelines:\\
1. The ``target" should introduce a contradiction when compared to the "source," without being a mere negation.\\
2. The ``sentence + hallucination" should be clearly distinguishable from the ``input sentence".\\
3. Your additions should be creative yet reasonable.\\
4. Avoid changing gender, color, or race of humans in the sentence.\\
5. The ``Correct Misalignment" should describe how the ``input sentence" diverges from the ``sentence + hallucination".\\\\

Input Sentence: A cola bottle is shown and then it is tossed\\
Sentence + Hallucination: A cola bottle is shown and then it is tossed along with a frisbee\\
Source: ``tossed"\\
Target: ``tossed along with a frisbee"\\
Correct Misalignment: There is no frisbee being tossed\\\\

Input Sentence: A person is playing a video game where they become aggressive towards a woman robot face\\
Sentence + Hallucination: A person is playing a video game where they become aggressive and release fireworks towards a woman robot face\\
Source: ``aggressive towards"\\
Target: ``aggressive and release fireworks towards"\\
Correct Misalignment: The person does not release fireworks at woman robot face\\\\

Input Sentence: A man is walking his dog\\
Sentence + Hallucination: A man is walking his dog while carrying a surfboard\\
Source: ``walking his dog"\\
Target: ``walking his dog while carrying a surfboard"\\
Correct Misalignment: The man does not carry a surfboard\\\\

Input Sentence: Children are playing in the park\\
Sentence + Hallucination: Children are playing in the park near a giant sculpture\\
Source: ``playing in the park"\\
Target: ``playing in the park near a giant sculpture"\\
Correct Misalignment: There is no giant sculpture in the park\\\\

Input Sentence: A woman is reading a book\\
Sentence + Hallucination: A woman is reading a book under a parasol\\
Source: ``reading a book"\\
Target: ``reading a book under a parasol"\\
Correct Misalignment: There is no parasol where the woman is reading a book\\\\

Remember: Only add elements; do not replace or remove any existing elements in the ``input sentence". Now it's your turn.\\\\

Input Sentence: $<$insert caption$>$\\
Sentence + Hallucination:\\
Source:\\
Target:\\
Correct Misalignment:\\
\bottomrule
\end{tabular} }
\caption{PaLM-2 LLM API prompt to generate contrast captions with \textit{Hallucination} misalignment.}
\label{tab:prompt_hallucination}
\end{figure*}

\begin{figure*}
\centering
\resizebox{\linewidth}{!}{

\begin{tabular}{p{1.3\linewidth}}
\toprule
Your objective is to generate a contradiction sentence using a provided ``input sentence" based on a specific ``misalignment scenario" called ``Event Misalignment". In this scenario, you should change the temporal order of the events in the sentence.\\\\

Your last task is to provide a ``Correct Misalignment" description, clarifying how the ``input sentence" is different from the ``sentence + event misalignment".\\\\

Key Requirements:\\
- The ``sentence + event misalignment" should be plausible and could theoretically occur in real life.\\
- Only focus on the temporal order; do not replace or remove any existing objects, actions or attributes in the ``input sentence".\\\\

Guidelines:\\
1. The ``sentence + event misalignment" should be clearly distinguishable from the ``input sentence".\\
2. Your changes should be creative yet reasonable.\\
3. Avoid changing gender, color, or race of humans in the sentence.\\
4. The ``Correct Misalignment" should describe how the ``input sentence" diverges from the ``sentence + event misalignment".\\\\

Input Sentence: A girl pretends to sneeze and drops something out of her hands and her friend starts to laugh and drops the phone\\
Sentence + Event Misalignment: A girl drops something out of her hands and then pretends to sneeze and her friend starts to laugh and drops the phone\\
Correct Misalignment: A girl first sneezes and then drops something out of her hands\\

Input Sentence: A boy is throwing a ball against a wall and a girl takes the ball and throws it.\\
Sentence + Event Misalignment: A girl takes the ball and throws it before the boy throws the ball against a wall\\
Correct Misalignment: A boy is throws the ball against the wall before the girl takes it and throws it\\\\

Input Sentence: A small crowd watches as a competitor performs a triple jump, then walks back to the starting mark.\\
Sentence + Event Misalignment: A small crowd watches a competitor walk to the starting mark, then perform a triple jump\\
Correct Misalignment: A competitor performs the triple jump before walking back to the starting mark\\\\

Input Sentence: A man wearing a black t-shirt is holding a cup of food in his right hand. He moves around a piece of food in his left hand to play with the ostrich.\\
Sentence + Event Misalignment: A man wearing a black t-shirt moves around a piece of food in his left hand to play with the ostrich before holding a cup of food in his right hand.\\
Correct Misalignment: A man is holding a cup of food before he moves around a piece of food to play with the ostrich\\\\

Input Sentence: A person is playing in the doorway, then they begin laughing and grab a doorknob and leave the room.\\
Sentence + Event Misalignment: A person is playing in the doorway, then they grab a doorknob and leave the room, and then they begin laughing.\\
Correct Misalignment: They begin laughing before they grabbed the doorknob and leave the room.\\\\

Now it's your turn.\\\\

Input Sentence: $<$insert caption$>$\\
Sentence + Event Misalignment:\\
Correct Misalignment:\\
\bottomrule
\end{tabular} }
\caption{PaLM-2 LLM API prompt to generate contrast captions with \textit{Event Order Flipping} misalignment.}
\label{tab:prompt_event}
\end{figure*}

\section{Human Annotation for Data Quality}
\label{sec:human_annotation_data_quality}

\begin{figure*}
    \centering
    \includegraphics[scale=0.35]{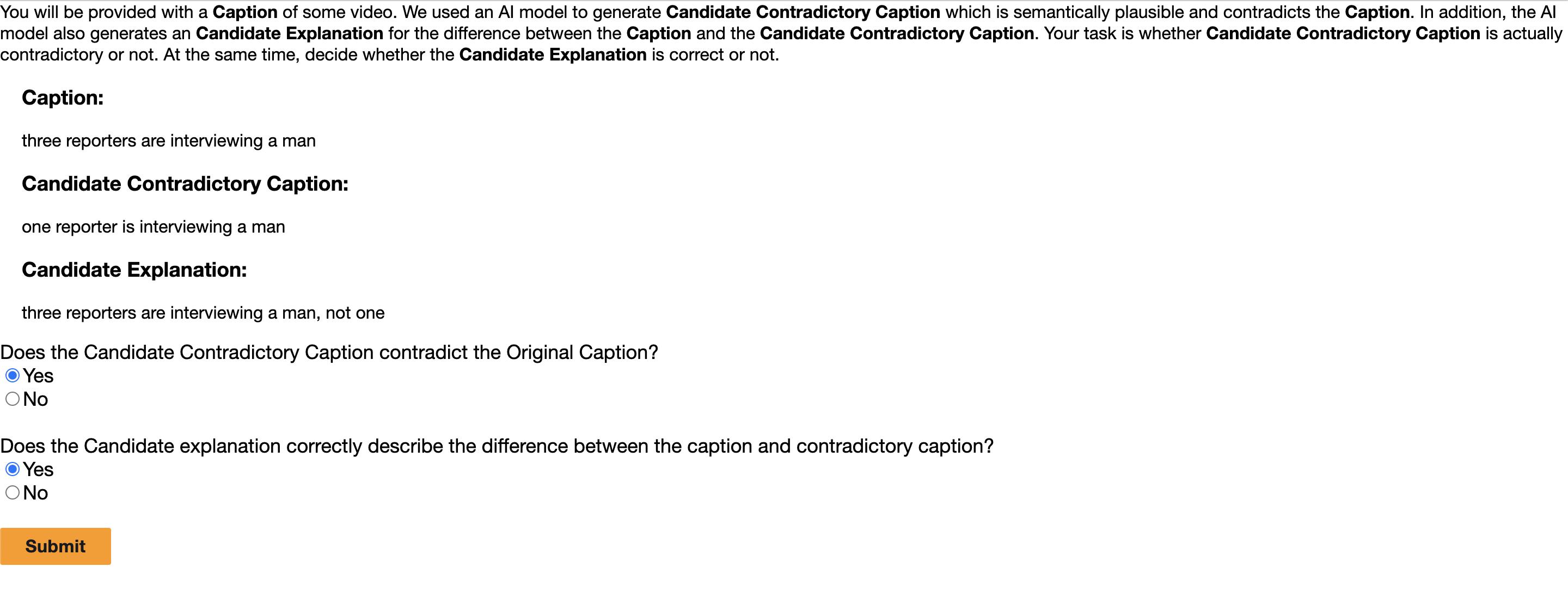}
    \caption{Screenshot of \name data quality assessment interface.}
    \label{fig:human_data_quality}
\end{figure*}

We use the workers from Amazon Mechanical Turk platform to assess the quality of the LLM generated data. We present the screenshot of the annotation interface in Figure \ref{fig:human_data_quality}. Specifically, the annotators are asked to decide whether the contrast captions contradict the original video captions. In addition, we ask the annotators to decide whether the generated natural language explanations correctly describe the discrepancy between the caption and contrast caption. The annotators are first asked to perform a qualification test and then selected for the final annotations. We assign one annotator per annotation instance. The human annotators were paid at $\$18$USD per hour, with the total expenditure of $\$180$ USD.

\section{\name (Human) Data Creation}
\label{sec:human_data_collection}

\begin{figure*}
    \centering
    \includegraphics[scale=0.35]{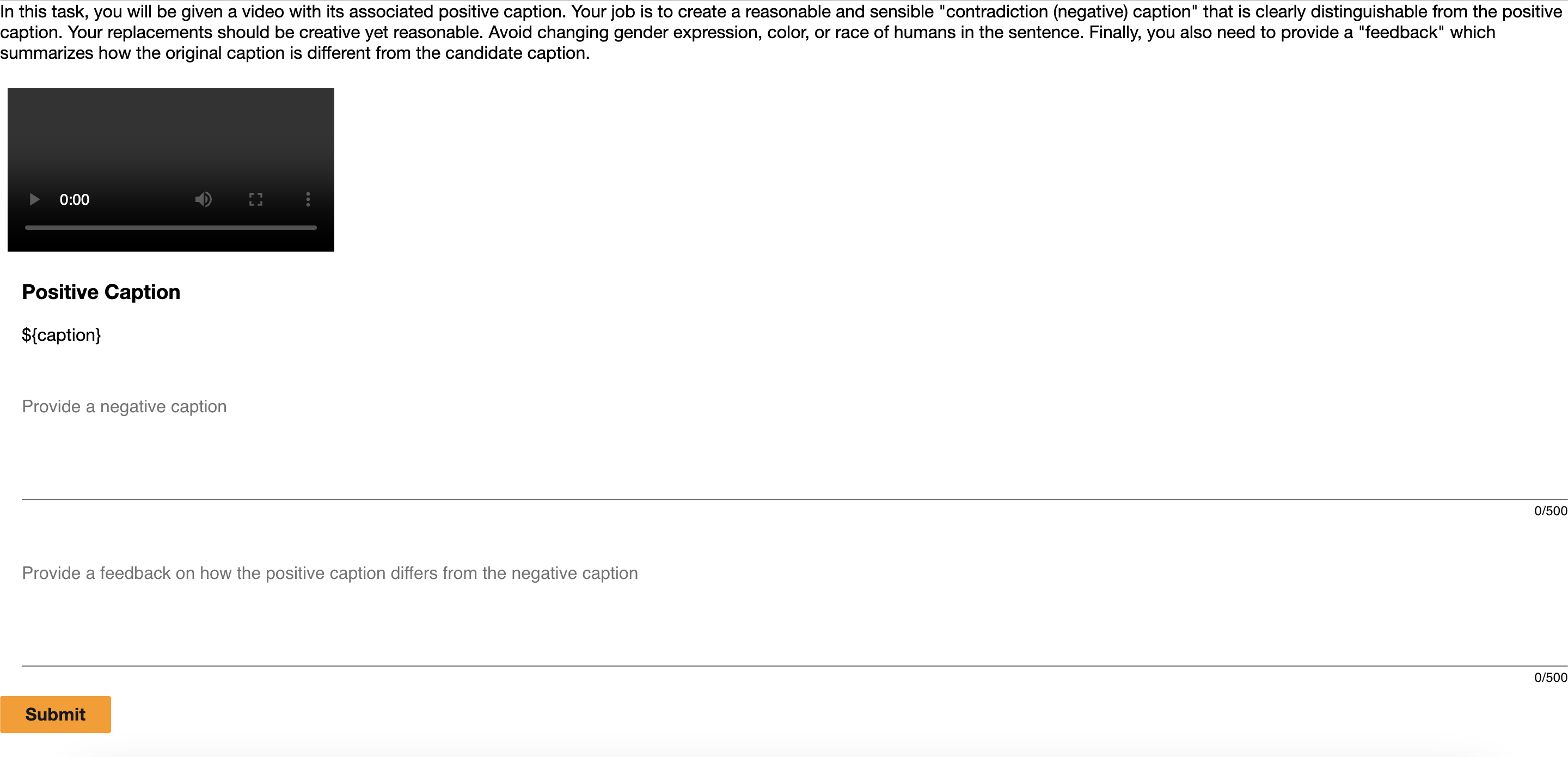}
    \caption{Screenshot of \name (Human) data collection interface.}
    \label{fig:human_data_creation}
\end{figure*}

\begin{figure*}
    \centering
    \includegraphics[scale=0.6]{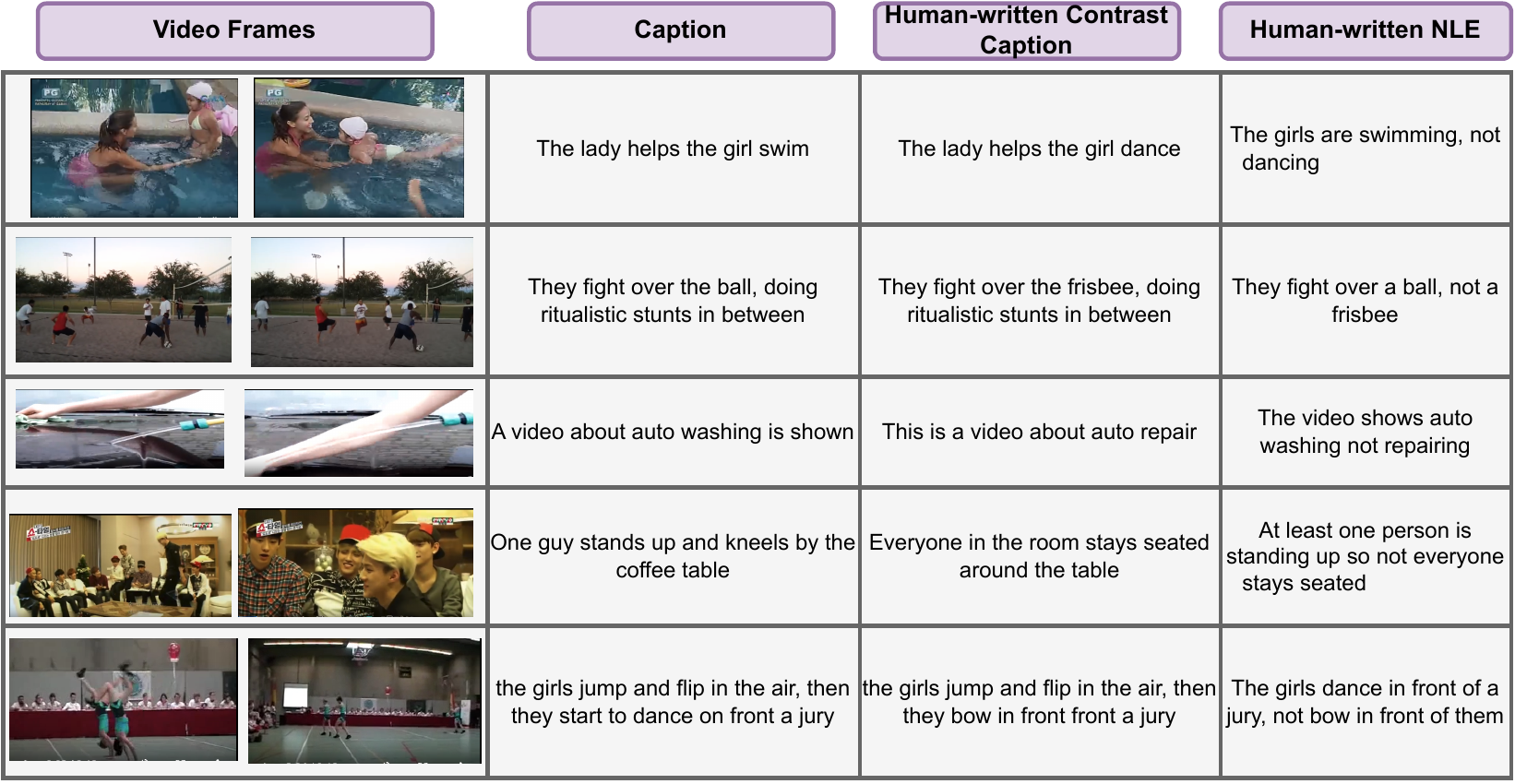}
    \caption{Example of the instances in the \name (Human) dataset.}
    \label{fig:videocon_human_examples}
\end{figure*}

To assess the generalization performance of our model, we create a human-written dataset in \name. Specifically, we ask the human annotators to create contrast captions and NLE while looking at the video segments taken from ActivityNet validation data \cite{caba2015activitynet} and their associated captions. We present the screenshot of the annotation interface in Figure \ref{fig:human_data_creation}. The annotators are \textbf{not} instructed to generate any specific kinds of misalignments in their contrast captions, and just asked generate semantically plausible contrast captions and their NLE. The annotators are first asked to perform a qualification test and then selected for the final annotations. We assign one worker per annotation instance. The human annotators were paid at $\$18$USD per hour, with the total expenditure of $\$260$ USD. We present a few examples from the \name (Human) dataset in Figure \ref{fig:videocon_human_examples}.

\section{Finetuning Details}
\label{sec:finetuning}

During finetuning, we use low-rank adaptation (LoRA) \cite{hu2021lora} of the \mplugowl (7B) \footnote{\url{https://github.com/X-PLUG/mPLUG-Owl/tree/main/mplug_owl_video}} applied to all the layers of the attention block i.e., query, key, value, output, gate, up, and down projection matrices. We set the LoRA $r = 32$, $\alpha = 32$, and dropout = $0.05$. The model is finetuned on the \name (LLM) training set (\S \ref{sec:llm_generated}) for 2 epochs. The finetuning was performed using Adam \cite{kingma2014adam} optimizer with the linear-warmup of $200$ steps followed by cosine decay learning schedule where the maximum learning rate = $10^{-4}$. We chose this learning rate after performing a hyperparameter search over $\{10^{-3}, 10^{-4}, 10^{-5}, 2 \times 10^{-5}\}$ based on the validation loss. We utilized $4$ A6000 GPUs with the total batch size of $64$ and one gradient accumulation step. We finetune our model by utilizing $32$ frames in the video. Specifically, we create $32$ segments of the video, and sample the middle frame from each video.

\section{Human Agreement for the Generated NLE Automatic Evaluation Methods}
\label{sec:human_agreement_nle}

Given the potential noise inherent in automated methods based on $Q^2$ and PaLM-2, we sought to ascertain their efficacy for NLE evaluation. We conducted a comparative analysis between these automated judgments and human judgments on a sample of $500$ instances derived from \name (LLM) and \name (Human), as shown in Table \ref{tab:validity}. We find that both the metrics achieve high ROC-AUC or agreement with the humans, thus, establishing their usefulness for scalable NLE evaluation.

\begin{table}[h]
\resizebox{\linewidth}{!}{%
\begin{tabular}{lcc}
\hline
                       & \textbf{\name (LLM)} & \textbf{\name (Human)} \\\hline
$Q^2$-Human ROC-AUC      & 92             & 89       \\     
PaLM-2-Human Agreement & 77.40\%        & 72.50\%          \\\hline 
\end{tabular}%
}
\caption{\small{Human agreement analysis to assess the efficacy of the $Q^2$ and PaLM-2 as entailment evaluators for NLE generation task. We find that both automatic metrics reliably estimate the human judgements for the task. Hence, both of them can be used for scalable NLE evaluation.}}
\label{tab:validity}
\end{table}

\section{Details about Downstream Tasks}
\label{sec:dets_downstream}

We provide details about the downstream task datasets and the evaluation setup in \S \ref{appen_sec:t2v} and \S \ref{appen_sec:videoqa}.

\subsection{Text to Video Retrieval}
\label{appen_sec:t2v}

We perform text-to-video retrieval evaluation on \textit{Something-Something} (SSv2) dataset \cite{goyal2017something,lei2022revealing} that covers a wide range of $174$ daily actions and around $100$K videos. Originally, the dataset captions are presented in two forms: \textit{label} and \textit{template}. In our work, we utilize \textit{SSv2-template} since it removes the bias in the evaluation due to object recognition instead of temporal modeling.

Following this, \cite{sevilla2021only} came up with a list of $18$ actions (classes) that require models to capture rich temporal-information in the video (e.g., `Moving away from [something] with your camera'). Each class contains $12$ videos associated with it. We call this dataset as \textbf{SSv2-Temporal} consisting of $216$ $(18 \times 12)$ candidate videos for every text query (action).

In addition, \cite{bagad2023test} create a subset called \textbf{SSv2-Events} with $49$ actions (classes) that consist two verbs in the action templates that are indicative of multiple events in the video (e.g., `Poking [something] so that it spins around'). Overall, this dataset consists $2888$ $(49 \times 12)$ candidate videos for every text query (action).

We use the video-text alignment models to rank each video for every action-specific text query. We report the mean average precision (mAP) performance of the models based on the ranking. We want a robust video-language model to achieve high mAP scores on this dataset.

\subsection{Video QA}
\label{appen_sec:videoqa}

We assess the VideoQA performance of the video-language alignment models on \textit{ATP-Hard} dataset \cite{buch2022revisiting}. It is a causal-temporal split \footnote{\url{https://stanfordvl.github.io/atp-revisit-video-lang//assets/atp-hard-ct4.txt}} of the Next-QA validation dataset \cite{xiao2021next} \footnote{\url{https://github.com/doc-doc/NExT-QA/blob/main/dataset/nextqa/val.csv}}. It consists of $2269$ instances $(V, Q, \{A_1, A_2, A_3, A_4, A_5\}, A)$ of video $V$, question $Q$, and five multiple-choice options $\{A_1, A_2, A_3, A_4, A_5\}$, and a ground-truth answer $A$. 

The aim of a video QA model is to choose the ground-truth answer from the multiple-choice options. To utilize a video-language alignment model for this task, we first recast the input $(Q, A_i)$ pairs into imperative statements using PaLM-2 LLM API. We present the LLM prompt in Figure \ref{fig:llm_prompt_vidqa}. For example, $Q = $ `what does the white dog do after going to the cushion?' and $A_i = $ `shake its body' is converted to a statement $S(Q, A_i) = $`The white dog shakes its body after going to the cushion'. We use the video-language alignment model to score $S(Q, A_i) \forall i \in \{1,2,3,4,5\}$. The statement with highest entailment score is considered as the model's prediction. We report the accuracy on the ATP-Hard dataset.

\begin{figure*}
\centering
\resizebox{\linewidth}{!}{

\begin{tabular}{p{1.3\linewidth}}
\toprule
You will be provided with a question along with the five multiple choice answers. You need to convert the question and every possible answer to an imperative statement. \\\\

Question: how do the two man play the instrument\\
Choices: \\
(A) roll the handle \\
(B) tap their feet\\
(C) strum the string\\
(D) hit with sticks\\
(E) pat with hand\\
Imperative Statements for every option:\\
(A) two man play the instrument by rolling the handle\\
(B) two man play the instrument by tapping their feet\\
(C) two man play the instrument by strumming the string\\
(D) two man play the instrument by hitting the sticks\\
(E) two man play the instrument by patting with hand\\\\

Question: how does the man cycling try to sell the watch to the man in the trishaw\\
Choices:\\
(A) give him catalogue\\
(B) show him a video\\
(C) show him the watch\\
(D) dismount his bicycle\\
(E) give him the watch strap\\
Imperative Statements for every option:\\
(A) The man cycling tries to sell the watch to the man in the trishaw by giving him the catalogue\\
(B) The man cycling tries to sell the watch to the man in the trishaw by showing him a video\\
(C) The man cycling tries to sell the watch to the man in the trishaw by showing him the watch\\
(D) The man cycling tries to sell the watch to the man in the trishaw by dismounting his bicycle\\
(E) The man cycling tries to sell the watch to the man in the trishaw by giving him the watch strap\\\\

Question: what does the white dog do after going to the cushion\\
Choices:\\
(A) drink again\\
(B) shake its body\\
(C) smells the black dog\\
(D) wagging tail\\
(E) touch lady in blue stripes\\
Imperative Statements for every option:\\
(A) white dog drinks again after going to the cushion\\
(B) white dog shakes its body after going to the cushion\\
(C) white dog smells the black dog after going to the cushion\\
(D) white dog wags its tail after going to the cushion\\
(E) white dog touches the lady in blue stripes after going to the cushion\\\\

Now it's your turn.\\\\

Question: Q\\
Choices:\\
(A) A1\\
(B) A2\\
(C) A3\\
(D) A4\\
(E) A5\\
Imperative Statements for every option:\\
\bottomrule
\end{tabular} }
\caption{Converting the QA pairs into imperative statements for VideoQA dataset.}
\label{fig:llm_prompt_vidqa}
\end{figure*}

% WARNING: do not forget to delete the supplementary pages from your submission 
% \input{sec/X_suppl}

\end{document}